\documentclass[10pt,twocolumn,letterpaper]{article}

\usepackage[pagenumbers]{wacv} % To force page numbers, e.g. for an arXiv version
\usepackage[accsupp]{axessibility}  % Improves PDF readability for those with disabilities.
\usepackage{multirow}
\usepackage{array}
\usepackage{siunitx}
\usepackage{bm}
\usepackage{algorithm}
\usepackage{algpseudocode}
\usepackage{listings}

\usepackage{tikz}
\usepackage[normalem]{ulem}
\usepackage{fontawesome5}
\usetikzlibrary{positioning, arrows.meta, calc, fit, shapes.geometric, backgrounds, decorations.pathreplacing, patterns}

\lstdefinestyle{augstyle}{
  basicstyle=\ttfamily\small,  % typewriter font
  breaklines=true,             % automatic line breaking
  breakatwhitespace=false,     % allow breaking inside long tokens
  xleftmargin=2em,             % indent the whole block
  columns=fullflexible,
  escapechar=@
}

\newcommand{\Ebase}{E_\text{base}}
\newcommand{\Dbase}{D_\text{base}}
\newcommand{\Etarget}{E_\text{target}}
\newcommand{\Dtarget}{D_\text{target}}
\newcommand{\Eft}{E_\text{adapted}}
\newcommand{\Dft}{D_\text{adapted}}

\newcommand{\noiselow}[1]{{\color{teal}#1}}
\newcommand{\noisehigh}[1]{{\color{orange}#1}}

\newcommand{\shead}[1]{#1}

\definecolor{wacvblue}{rgb}{0.21,0.49,0.74}
\usepackage[pagebackref,breaklinks,colorlinks,allcolors=wacvblue]{hyperref}

\def\wacvPaperID{475} % *** Enter the WACV Paper ID here
\def\confName{WACV}
\def\confYear{2027}

\title{Adaptable Watermarking: Transforming Image Watermarking Models via Lightweight Adapters%
\thanks{Accepted to the IEEE/CVF Winter Conference on Applications of Computer Vision (WACV) 2027.}}
\renewcommand{\thetitle}{Adaptable Watermarking: Transforming Image Watermarking Models via Lightweight Adapters}

\author{Mehrdad Saberi$^{1}$ \\
\and Vishal Asnani$^{2}$ \\
\and Shruti Agarwal$^{2}$ \\
\and Soheil Feizi$^{1}$ \\
\and John Collomosse$^{2,3}$\\
\and \vspace{-1.1cm}\\
$^{1}$University of Maryland.\\
$^{2}$Adobe Research.\\
$^{3}$DECaDE, University of Surrey.\\
{\tt\small \{msaberi, sfeizi\}@umd.edu ~~~\{vasnani, shragarw, collomos\}@adobe.com}}

\begin{document}
\maketitle

\begin{abstract}

We explore the adaptability of neural image watermarking models; the extent to which one watermarking method can be transformed into another through lightweight adaptation. We introduce a framework that enables encoder-decoder watermarking architectures to become modular and upgradable, supporting multiple watermarking styles without re-training full models. Adaptation is performed by attaching a compact parameter-efficient adapter and fine-tuning under a multi-domain residual alignment loss (spatial, gradient, and frequency) while preserving bit accuracy via a secret-loss term.  We evaluate this approach across a suite of recent watermarking models, including TrustMark, InvisMark, RoSteALS, HiDDeN, and WAM, demonstrating that adapted encoders closely reproduce target residuals and that adapted decoders recover messages with near-optimal accuracy under standard augmentations and distortions.  We show that cross-model adaptation, with applications in interoperable and upgradable watermarking, can be achieved with as few as 1-2M trainable parameters per component (e.g., the 2M decoder adapter used in our multi-decoder inference pipeline).
  
\end{abstract}
\section{Introduction}
\label{sec:intro}

\begin{figure*}[t]
\centering
\includegraphics[width=\textwidth]{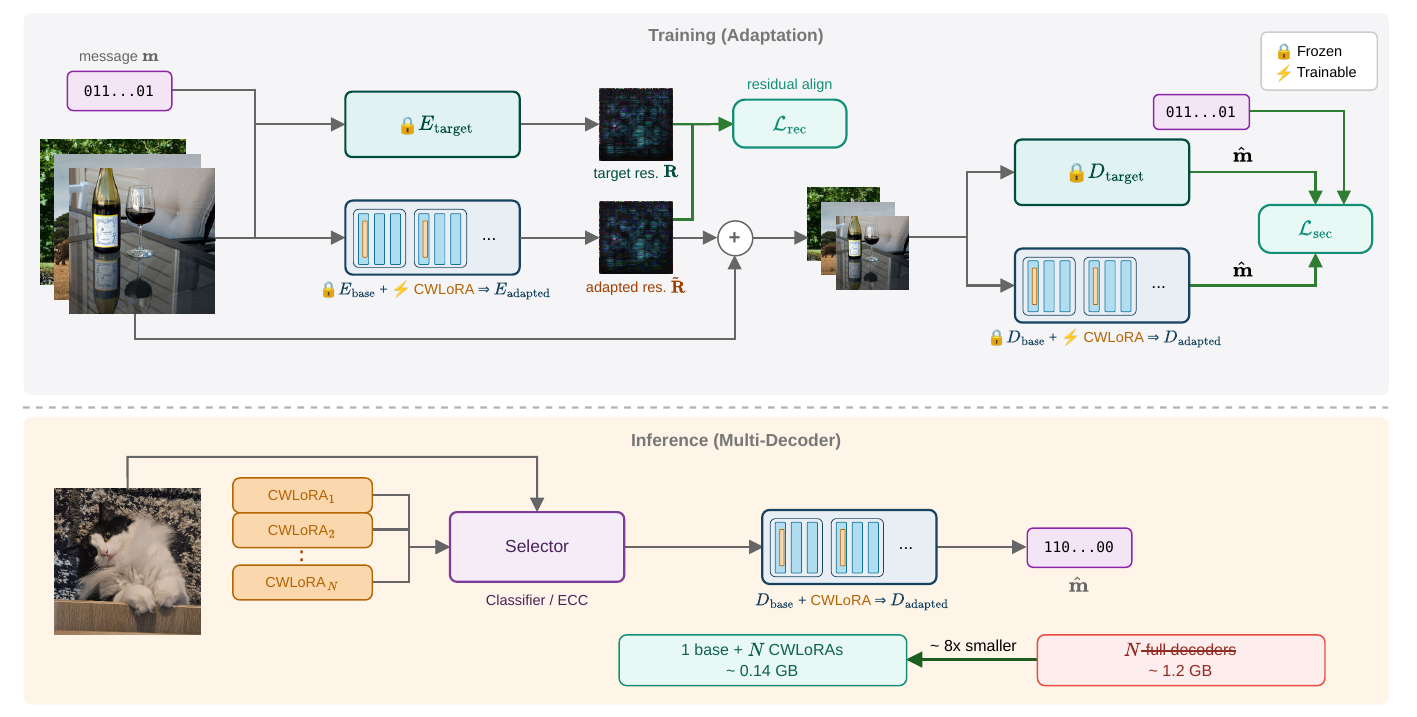}
\caption{\textbf{Method overview.} \textbf{Top (Training):} a frozen base watermarking pair $(\Ebase,\Dbase)$ is augmented with compact trainable CWLoRA adapters ($\sim$1--2M params, shown in orange inside the conv blocks) to produce $(\Eft,\Dft)$ that mimic a frozen target pair $(\Etarget,\Dtarget)$. Only the CWLoRA adapters are trained, via a residual alignment loss $\mathcal{L}_\text{rec}$ (matching $\bm{\tilde{R}}$ to the target residual $\bm{R}$) and a BCE secret loss $\mathcal{L}_\text{sec}$ (recovering $\bm{m}$ after the target noise module $\mathcal{A}_\text{tgt}$). \textbf{Bottom (Inference):} a single frozen $\Dbase$ and a stack of $\sim$2M-parameter CWLoRA adapters replace $N$ full decoders; a selector (classifier- or ECC-based) picks the right CWLoRA for a query image, reducing loaded decoder size from $\sim$1.2\,GB to $\sim$0.14\,GB ($\sim8\times$) while preserving near-oracle accuracy.}
\label{fig:method_overview}
\end{figure*}

Invisible image watermarking embeds machine-readable signals into images while preserving human visual quality and robustness to common post-processing. It underpins copyright protection, tamper detection, and, increasingly, provenance tracking for AI-generated content \cite{Balan-WMF-2023,Collomosse-IEEECGA-2026}. Decades of work have produced a broad toolkit, including classical transform-domain schemes \cite{cox1997spread,alhaj2007dwtdct}, modern in-generation watermarks for diffusion models \cite{yang2024gaussianshading,gunn2025prc,wen2023treering}, and watermarking for autoregressive generators \cite{tong2025indexmark, jovanovic2025wmar}. Surveys highlight that deep learning has become a dominant paradigm across these directions \cite{hosny2024survey,zhang2021survey}.

In this paper, we focus on neural encoder-decoder watermarking for images \cite{zhu2018hidden,hosny2024survey,wang2021survey}, which are popular for their ability to watermark arbitrary inputs, high robustness, and imperceptibility. To our knowledge, this is the first work to explore neural watermarking methods using parameter-efficient adapters. Numerous architectures have been proposed, including CNN-based models such as TrustMark \cite{bui2023trustmark}, HiDDeN \cite{zhu2018hidden}, and StegaStamp \cite{tancik2020stegastamp}, latent-space designs like RoSteALS \cite{bui2023rosteals}, and transformer-based or segmentation-aware models such as WAM \cite{sander2024watermark}. Although these methods share a common abstraction, they are typically trained and deployed as separate, monolithic models with different residual patterns, visual impact, and robustness profiles.

This fragmentation is problematic for real systems that must support multiple watermark styles or evolve them over time. A platform or model provider may wish to (i) upgrade an existing watermark to a new design, (ii) interoperate with third-party watermarking schemes, or (iii) offer multiple watermark ``profiles'' (e.g., stronger robustness vs.\ higher visual fidelity) without maintaining separate large models for each. Such changes usually require retraining and redeploying entire encoder-decoder pairs, which can be expensive and brittle \cite{hosny2024survey,wang2021survey}. In production, however, shipping $N$ separate models is costly; a modular framework that supports multiple profiles from a single frozen backbone drastically reduces model loading time and client download sizes, enabling efficient updates without infrastructure overhauls.

We study this idea through the lens of \emph{adaptability} between watermarking models (Figure~\ref{fig:method_overview}). Let $(\Ebase,\Dbase)$ be a base encoder and decoder and $(\Etarget,\Dtarget)$ a target watermarking pair from another method (e.g., TrustMark P/Q, InvisMark, RoSteALS, HiDDeN, or WAM) \cite{bui2023trustmark,xu2025invismark,bui2023rosteals,zhu2018hidden,sander2024watermark}. We insert a compact adapter into the base models and fine-tune only this adapter, obtaining adapted models $(\Eft,\Dft)$ that aim to mimic the behavior of $(\Etarget,\Dtarget)$ on natural images and binary messages. Rather than matching parameters, we define similarity through behavior: encoder adaptation encourages the watermark residuals produced by $\Eft$ to match those of $\Etarget$ and to remain decodable by $\Dtarget$ under standard distortions; decoder adaptation encourages $\Dft$ to reliably decode messages embedded by either $\Etarget$ or $\Eft$.

Concretely, we guide adaptation with a novel multi-domain watermark residual alignment loss and a standard BCE secret loss. We use Convolutional-Weight LoRA (CWLoRA), which applies low-rank updates directly to convolutional kernels~\cite{lora}. Unlike common post-layer convolutional adapters~\cite{peft,yeh2024navigating,ryu2022diffusionlora}, CWLoRA merges into the base weights with no inference-time overhead, and gives the best accuracy/size trade-off in our benchmarks; at rank 16 it needs only 0.9M (encoder) and 2.0M (decoder) parameters, roughly 10\% of a typical base model.

% To realize this, we introduce a novel multi-domain watermark residual alignment loss and also use a BCE-based secret loss. We implement adaptation using our novel Convolutional-Weight LoRA (CWLoRA), a parameter-efficient convolutional weight-space adapter. At rank 16, it requires only 0.9M parameters for the encoder and 2.0M for the decoder, roughly 10\% of a typical base model and a negligible fraction of larger targets.

% Across our experiments, we show that starting from a convolutional TrustMark encoder and from multiple base decoders (including those of TrustMark, InvisMark, and RoSteALS), CWLoRA adapters can transform the base systems to closely reproduce the residual patterns and decoding behavior of diverse target watermarking methods. Adapted encoders produce residuals well aligned with their targets (Figure~\ref{fig:residual_grid}) and retain high message recovery rates under common augmentations, while adapted decoders achieve competitive, often near-target robustness relative to the original target decoders, even when architectures differ substantially. Importantly, the adapted models remain \emph{interoperable} with the original target models: the target decoder can read messages from the adapted encoder and vice versa, confirming behavioral fidelity beyond residual alignment.
Across our experiments, CWLoRA adapters closely reproduce target residual patterns (Figure~\ref{fig:residual_grid}) and near-target decoding even across architectures, and the adapted models remain \emph{interoperable} with the originals: the target decoder can read messages from the adapted encoder and vice versa.

Beyond per-method adaptation, we also build a \emph{multi-decoder} inference pipeline that replaces full decoders with lightweight adapters. Using either an ECC-based (Error Correction Codes) plug-and-play decoder selector or a small classifier-based selector, the adapterized system attains near-oracle decoding accuracy while shrinking total loaded decoder size from $\sim$1.2\,GB to $\sim$0.14\,GB. With the classifier-based selector, adapterized decoders reach $96.9\%$ (clean) and $85.7\%$ (noised), matching/exceeding the original-decoder setting; the ECC-based selector is within $\approx$0.6--1.0\% of the originals (Table~\ref{tab:multi-dec}).

\begin{figure*}[t]
    \centering
    \includegraphics[width=1.0\linewidth]{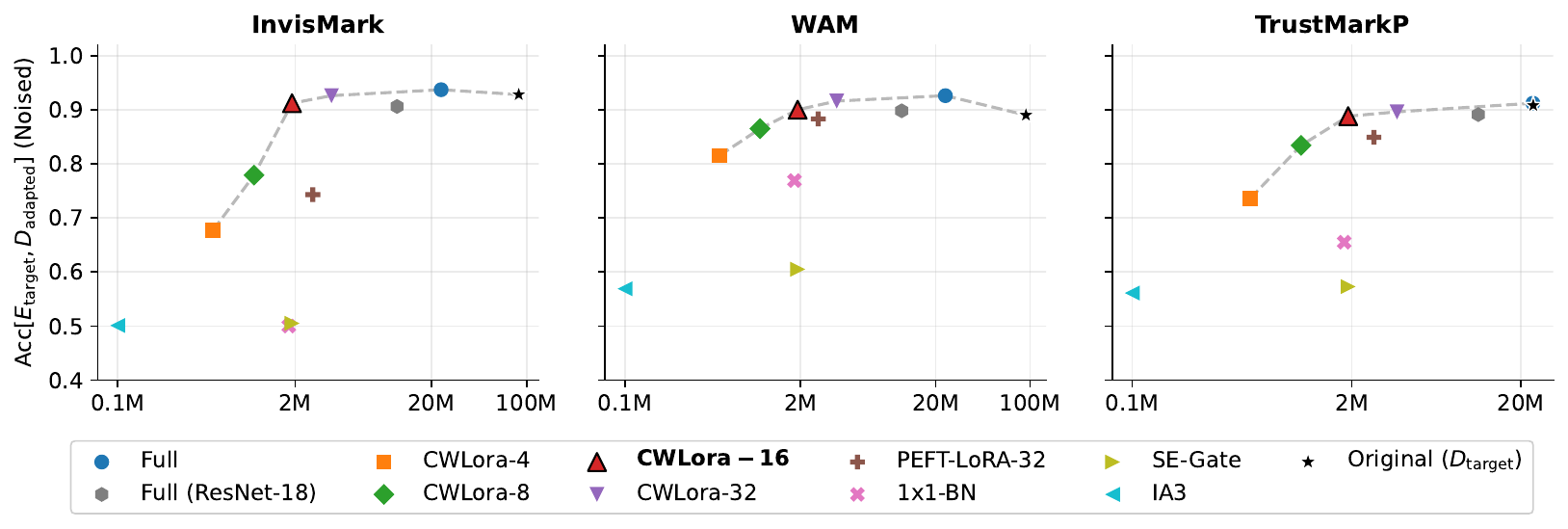}
    \caption{\textbf{Adapter comparison for decoder-only adaptation.} Bit accuracy under noise vs. number of parameters for decoder adaptation with different types of adapters. The base decoder is from the TrustMarkQ method.}
    \label{fig:adapter_comparison}
    \vspace{-0.2cm}
\end{figure*}

We make the following technical contributions:

\begin{itemize}
    \item We formalize cross-method adaptation for neural watermarking via lightweight adapters, proposing objectives for residual alignment and decoder compatibility.
    % \item We introduce CWLoRA, a convolutional weight-space adapter for parameter-efficient fine-tuning that introduces no inference-time overhead.
    \item We apply Convolutional-Weight LoRA (CWLoRA), a weight-space low-rank adapter for convolutional backbones built on LoRA~\cite{lora}, and show it outperforms common post-layer convolutional adapters on this task while introducing no inference-time overhead.
    \item We demonstrate effective encoder and decoder adaptation across diverse target watermarks, achieving aligned residuals and near-target decoding despite large architectural differences, and show that these results hold across multiple base methods, out-of-distribution perturbations, and cross-dataset evaluations.
    \item We show that CWLoRA on a strong base preserves perceptual quality that full fine-tuning of a comparably small model cannot achieve, confirming that the base's pre-existing priors are essential.
    \item We build an adapterized multi-decoder inference pipeline with ECC and classifier-based selection, yielding near-oracle accuracy while reducing the footprint by $\sim8\times$, enabling flexible deployment from a single base model.
\end{itemize}

% \ms{add numeric details of robustness and psnr}

% \ms{add multi-decoder inference}

% \ms{add final contributions}

\section{Background and Related Work}
\label{sec:background}

\noindent \textbf{Image Watermarking.}
Traditional image watermarking methods operate in spatial or transform domains via spread-spectrum and DCT/DWT-based embedding~\cite{cox1997spread,alhaj2007dwtdct}. Most modern methods adopt a neural encoder-decoder trained end-to-end with robustness-oriented noise layers~\cite{zhu2018hidden,jia2021mbrs}, sometimes combined with invertible networks for multi-image hiding~\cite{guan2022deepmih}. Representative encoder-decoder image schemes include HiDDeN~\cite{zhu2018hidden}, StegaStamp~\cite{tancik2020stegastamp}, RivaGAN~\cite{zhang2019rivagan}, MBRS~\cite{jia2021mbrs}, and DeepMIH~\cite{guan2022deepmih}, while more recent architectures target arbitrary resolutions and high robustness, such as TrustMark~\cite{bui2023trustmark}, RoSteALS~\cite{bui2023rosteals}, InvisMark~\cite{xu2025invismark}, SynthID \cite{gowal2025synthid}, and the localized WAM framework~\cite{sander2024watermark}. In parallel, in-generation watermarking for generative models operates inside diffusion or autoregressive pipelines~\cite{wen2023treering,yang2024gaussianshading,gunn2025prc,tong2025indexmark,jovanovic2025wmar}. Complementary methods such as JIGMark focus on enhancing robustness against specific modifications such as diffusion-based image edits~\cite{pan2024jigmark}. Together, these lines of work highlight the diversity of architectures and training recipes, and the lack of a unifying mechanism to make different watermarking styles interoperable at the model level.

\noindent \textbf{Watermark Adaptation.}
We use the term \emph{watermark adaptation} to refer to attempts to make watermarking systems more unified and interoperable across methods, enabling a deployed watermarking stack to accommodate, modify, or switch between different watermark designs without retraining full models.
% Conceptually, such adaptation would allow a single framework to support multiple watermark ``styles,'' evolve them over time, and interoperate across providers via controlled modifications of encoder-decoder behavior rather than entirely separate pipelines.

A related but adversarial line of work considers \emph{forgery} attacks that attempt to mimic the patterns of target watermarks, usually with limited model access \cite{saberi2024robustness, yang2024simple, soucek2025transferable}. \cite{petrov2025coexistence} studies the \emph{coexistence} and ensembling of multiple deep watermarks applied to the same image.
Unlike watermark co-existence methods~\cite{petrov2025coexistence}, our goal is not to layer multiple methods but to transform one method into another. TrustMark \cite{bui2023trustmark} introduces a dedicated watermark-removal network to erase an existing watermark and re-embed a new one, providing a practical mechanism for re-watermarking and upgrade of identifiers at the image level.

\section{Methodology}
\label{sec:methodology}

\begin{table*}[t]
\centering
\caption{\textbf{Decoder-only adaptation across base decoders.} The “Original” row reports the baseline \(\mathrm{Acc}[\Etarget,\Dtarget]\) (clean/noised). Subsequent rows show \(\mathrm{Acc}[\Etarget,\Dft]\) after adapting with different base decoders \(\Dbase\) using CWLoRA-16. The bottom row lists the \(\mathrm{PSNR}\) of the original target encoder, which can be a proxy for difficulty of decoder adaptation.
}
\label{tab:base-decoder-comparison}
\renewcommand{\arraystretch}{1.15}
%\resizebox{\textwidth}{!}{%
\begin{tabular}{l c c c c c c c c c c}
\toprule
\multirow{2}{*}{Base Decoder} &
\multicolumn{2}{c}{InvisMark} &
\multicolumn{2}{c}{RoSteALS} &
\multicolumn{2}{c}{TrustMarkQ} &
\multicolumn{2}{c}{TrustMarkP} &
\multicolumn{2}{c}{WAM} \\
\cmidrule(lr){2-3}\cmidrule(lr){4-5}\cmidrule(lr){6-7}\cmidrule(lr){8-9}\cmidrule(lr){10-11}
& \shead{Clean} & \shead{Noised}
& \shead{Clean} & \shead{Noised}
& \shead{Clean} & \shead{Noised}
& \shead{Clean} & \shead{Noised}
& \shead{Clean} & \shead{Noised} \\
\midrule
Original ($\Dtarget$)   &    0.954 &   0.928 &  0.995 & 0.969 & 1.000 & 0.958 & 0.995 & 0.908  & 1.000 & 0.890 \\
\midrule
InvisMark & -- & -- & 0.994 & 0.959 &  0.988 & 0.910 & 0.906 & 0.731 & 0.990 & 0.925\\
RoSteALS & 0.937 & 0.881 & -- & -- & 0.995 & 0.897 & 0.981 & 0.842 &  0.974 & 0.917\\ 
% TrustMarkQ$_{high\_noise}$  & 0.9533 & 0.8780 & 0.9947 & 0.9525 & 0.9948 & 0.9848 & 0.9827 & 0.9310\\
TrustMarkQ  & 0.959 & 0.909 & 0.995 & 0.967 & -- & -- & 0.991 & 0.885 & 0.983 & 0.905\\
% TrustMarkQ$_{low\_noise}$   & 0.9518 & 0.8713 & 0.9913 & 0.9469  & 0.9944 & 0.9843 & 0.9894 & 0.9413 \\
% TrustMark--Q$_{no\_noise}$   & 0.4998 & 0.4998 & 0.9942 & 0.9833 & 0.9870 & 0.9413 & & &  0.9890 & 0.9376  \\
% Imagenet Pretrained    & 0.5000 & 0.5000 & 0.5002 & 0.5002 & 0.9941 & 0.9837 & 0.9895 & 0.9478 \\
ResNet-50 ImageNet    & 0.500 & 0.500 & 0.994 & 0.965 & 0.995 & 0.931 & 0.501 & 0.501   & 0.990 & 0.925 \\
\midrule \midrule
\(\mathrm{PSNR}[\Etarget]\)  & \multicolumn{2}{c}{47.37} & \multicolumn{2}{c}{28.42} &  \multicolumn{2}{c}{39.69}  & \multicolumn{2}{c}{48.33} & \multicolumn{2}{c}{36.96} \\
\bottomrule
\end{tabular}%
%}
\vspace{-0.4cm}
\end{table*}

\subsection{Terminology and Setup}
We measure the \emph{adaptability} of a \emph{base} watermarking method to a \emph{target} method. Let the encoder/decoder of the base method be \(\Ebase\) and \(\Dbase\), and those of the target method be \(\Etarget\) and \(\Dtarget\).
Applying a lightweight adapter and fine-tuning the base models to mimic the behavior of the target models yields adapted variants \(\Eft\) and \(\Dft\). We discuss the adapter options and introduce the CWLoRA adapter in Section~\ref{sec:method_cwlora}.

Let \(\bm{X}\sim\bm{\mathcal{X}}\) denote a (cover) image and \(\bm{m}\in\{0,1\}^N\) denote an \(N\)-bit message.
We use the following losses:
\begin{itemize}
  \item \textbf{Reconstruction loss} \(\mathcal{L}_\text{rec}[E_1,E_2]\): a similarity loss between two encoders \(E_1\) and \(E_2\), measuring how closely their watermarked outputs (or residuals) match on the same \((\bm{X},\bm{m})\). Intuitively, it quantifies how well one encoder reproduces the watermarking behavior of another.
  \item \textbf{Secret loss} \(\mathcal{L}_\text{sec}[E,D]\): a decoding loss between an encoder \(E\) and a decoder \(D\), measuring how accurately \(D\) can recover the embedded message \(\bm{m}\) from images produced by \(E\), possibly after stochastic image degradations.
  % In the next subsections we instantiate this loss using a binary cross-entropy objective and a noise/augmentation module.
\end{itemize}

\subsection{Decoder-only Adaptation}
\label{sec:method_deconly}
We begin the analysis with a simpler adaptation setting, which involves a single objective. Here, we \emph{freeze} the target encoder \(\Etarget\) and fine-tune a lightweight adapter module attached to the decoder \(\Dbase\), such that the final decoder \(\Dft\) can decode messages produced by \(\Etarget\). The adapted encoder \(\Eft\) is not involved in this setting.
Formally, we minimize a secret loss between the target encoder and the adapted decoder.

\vspace{0.2cm}
\noindent \textbf{Noise module and secret loss.}
Whenever a watermarked image is decoded, we pass it through a stochastic \emph{noise/augmentation module} that simulates the distortions the watermark is expected to withstand. We use the target watermark's training noise module, denoted \(\mathcal{A}_\text{tgt}\), which applies random image degradations such as JPEG compression, blurring, perspective changes, and random crops.
This module is fixed and shared across training and evaluation to ensure fair comparison across adapted models.

For encoder \(E\) and decoder \(D\), secret loss is defined as
\begin{equation}
\label{eq:sec-loss-definition}
\mathcal{L}_\text{sec}[E,D]\triangleq \mathbb{E}_{\bm{X},\bm{m},\mathcal{A}\sim\mathcal{A}_\text{tgt}}\!\left[
\mathrm{BCE}\Big(D\big(\mathcal{A}(E(\bm{X},\bm{m}))\big),\,\bm{m}\Big)
\right],
\end{equation}
where \(\mathrm{BCE}\) is the bit-wise binary cross-entropy between the decoder's logits and the ground-truth message, and the expectation is over the stochasticity of the noise/augmentation pipeline \(\mathcal{A}_\text{tgt}\).

In decoder-only adaptation, we instantiate this loss with the frozen target encoder and the trainable decoder:
\begin{equation}
\label{eq:dec-standalone}
\min_{\Dft}\;\;
\mathcal{L}_\text{sec}[\Etarget,\Dft],
\end{equation}
using the same \(\mathcal{A}_\text{tgt}\) during training. In Section~\ref{sec:exp_deconly}, we present decoder adaptation results (bit accuracy in both clean and noised settings) across multiple base-target method pairs and adapter configurations.

% \paragraph{Decoder-only evaluation.}
% To obtain a fair evaluation of decoder-only adaptation, we also evaluate exclusively under the target watermark's noise module \(\mathcal{A}_\text{tgt}\).
% We report (i) \emph{clean} bit accuracy, where messages embedded by \(\Etarget\) are decoded by \(\Dft\) without additional corruption, and (ii) \emph{noised} bit accuracy, where the same messages are decoded after passing through stochastic degradations drawn from \(\mathcal{A}_\text{tgt}\).
% These two metrics quantify how well the adapted decoder \(\Dft\) has learned to decode the target encoder under both ideal and realistic (noisy) conditions.

\subsection{Encoder-Decoder Adaptation}
\label{sec:method_encdec}

We next consider a full adaptation of watermark models, where \(\Eft\) and \(\Dft\) are trained jointly.
In this setting, \(\Eft\) learns to match \(\Etarget\) while staying decodable by \(\Dtarget\), and \(\Dft\) learns to decode messages from both \(\Eft\) and \(\Etarget\).
All secret-loss terms reuse the same definition in~\eqref{eq:sec-loss-definition} and the same target noise module \(\mathcal{A}_\text{tgt}\) as in the decoder-only setup, ensuring that training and evaluation are aligned.

\vspace{0.2cm}
\noindent \textbf{Reconstruction loss.}
To measure encoder similarity more directly, we compare their residuals.
Given an image-message pair \((\bm{X},\bm{m})\), we define
\[
\bm{R} \;=\; \Etarget(\bm{X},\bm{m}) - \bm{X}, \qquad
\bm{\tilde{R}} \;=\; \Eft(\bm{X},\bm{m}) - \bm{X},
\]
and use a multi-term residual-matching objective:
\begin{equation}
\label{eq:rec-loss}
\begin{aligned}
\mathcal{L}_{\text{rec}}
= {}& \alpha\,\Phi_\epsilon\!\big(\bm{\tilde{R}}-\bm{R}\big)
     + \beta\,\Phi_\epsilon\!\big(\nabla_S \bm{\tilde{R}}-\nabla_S \bm{R}\big) \\
   & \quad + \gamma\,\Phi_\epsilon\!\big(\,|\bm{\mathcal{F}}\bm{\tilde{R}}|-|\bm{\mathcal{F}}\bm{R}|\,\big)
     + \delta\,\big(1-\cos(\bm{\tilde{R}},\bm{R})\big),
\end{aligned}
\end{equation}
where \(\Phi_\epsilon(\bm{Z})=\sum_{z \in \bm{Z}} \sqrt{z^2+\epsilon^2}\) is the element-wise Charbonnier penalty summed over all entries (pixels and channels), \(\nabla_S\) is a per-channel Sobel operator, \(\bm{\mathcal{F}}\) is the 2D discrete Fourier transform and \(|\cdot|\) denotes magnitude, and
\(\cos(\bm{\tilde{R}},\bm{R})=\frac{\langle \mathrm{vec}(\bm{\tilde{R}}),\,\mathrm{vec}(\bm{R})\rangle}{\|\bm{\tilde{R}}\|_2\,\|\bm{R}\|_2+\tau}\) is the cosine similarity of vectorized residuals.
The weights \(\alpha,\beta,\gamma,\delta \allowbreak \ge 0\) control the relative contribution of each term. In Appendix~\ref{sec:res_hparam_analysis}, we demonstrate results on how the proposed residual-matching loss yields improved encoder adaptation.

% \ms{add an ablation study on different reconstruction losses and compare final PSNR metric}

\vspace{0.2cm}
\noindent \textbf{Encoder objective.}
The encoder \(\Eft\) is optimized with a combination of reconstruction and secret losses:
\begin{equation}
\label{eq:enc-loss-definition}
\begin{aligned}
\min_{\Eft}\;\;
\mu_1\, & \mathcal{L}_\text{rec}[\Eft,\Etarget]
\\
+ & \ \mu_2\,\mathcal{L}_\text{sec}[\Eft,\Dtarget],
\end{aligned}
\end{equation}
where \(\mathcal{L}_\text{sec}\) is the BCE-based secret loss defined in~\eqref{eq:sec-loss-definition} with the target noise module \(\mathcal{A}_\text{tgt}\).
This objective explicitly aligns \(\Eft\) with \(\Etarget\) in residual space while enforcing the target decoder \(\Dtarget\) to reliably recover the embedded messages from \(\Eft(\bm{X},\bm{m})\).
We omit \(\mathcal{L}_\text{sec}[\Eft,\Dft]\) from the encoder objective to isolate reconstruction alignment; $\Dft$ is not optimized on this branch.

\vspace{0.2cm}
\noindent \textbf{Decoder objective.}
On the decoder side, we simultaneously encourage compatibility with both the adapted encoder and the target encoder, which is the difference compared to the \emph{decoder-only} adaptation.
Using the same secret loss \(\mathcal{L}_\text{sec}\) and noise module \(\mathcal{A}_\text{tgt}\) as above, we minimize
\begin{equation}
\label{eq:dec-paired}
\begin{aligned}
\min_{\Dft}\;\; 
\lambda_1\, & \mathcal{L}_\text{sec}[\Eft,\Dft]
\\
+ & \ \lambda_2\,\mathcal{L}_\text{sec}[\Etarget,\Dft],
\end{aligned}
\end{equation}
with \(\lambda_1,\lambda_2\ge 0\) controlling the trade-off between decoding from the adapted and target encoders.

\subsection{CWLoRA Adapter}
\label{sec:method_cwlora}

To enable parameter-efficient adaptation, we attach lightweight adapters to the base models \((\Ebase,\Dbase)\).
Because most watermarking encoders/decoders in our study are convolutional (the main exception is WAM’s ViT decoder), we focus on CNN-suitable adapters.
We benchmark several alternatives, including IA$^3$ \cite{ia3}, a $1{\times}1$ bottleneck \cite{1x1-bn}, and the convolutional PEFT-LoRA implementation from Hugging Face~\cite{peft} that adds post-layer $1{\times}1$ and \(K{\times}K\) convolutions (details in Appendix~\ref{sec:adapter_details}; see Fig.~\ref{fig:adapter_comparison}).

While low-rank weight-space updates are well established for linear layers~\cite{lora}, common convolutional adapters (e.g., PEFT-LoRA) typically inject post-layer additive convolutions instead~\cite{peft,yeh2024navigating,ryu2022diffusionlora}. We propose applying the former directly to convolutional kernels; we call this \emph{Convolutional-Weight LoRA} (CWLoRA). As we show in Section~\ref{sec:exp_deconly}, this design is better suited to watermark adaptation than the post-layer variant.

A CWLoRA-$r$ adapter applies direct sparse updates with rank $r$ on the weights of convolutional layers. Formally, for each layer $\ell$ with kernel
$W^{(\ell)} \in \mathbb{R}^{C^{(\ell)}_{\text{out}} \times C^{(\ell)}_{\text{in}} \times K^{(\ell)} \times K^{(\ell)}}$,
the CWLoRA-$r$ adapter introduces a low-rank update $\Delta W^{(\ell)}$ parameterized by
\[
A^{(\ell)} \in \mathbb{R}^{r \times C^{(\ell)}_{\text{in}} \times K^{(\ell)} \times K^{(\ell)}}, \quad
B^{(\ell)} \in \mathbb{R}^{r \times C^{(\ell)}_{\text{out}} \times K^{(\ell)} \times K^{(\ell)}},
\]
and defines the rank-$r$ update via an elementwise spatial/channel contraction:
\begin{equation*}
\begin{aligned}
\Delta W^{(\ell)}_{o,i,u,v}
&= \alpha_{\text{cw}} \sum_{t=1}^{r} B^{(\ell)}_{t,o,u,v}\, A^{(\ell)}_{t,i,u,v}, \\
\tilde{W}^{(\ell)} &= W^{(\ell)} + \Delta W^{(\ell)}.
\end{aligned}
\end{equation*}
Here $\alpha_{\text{cw}}$ is a scaling constant (set to $\alpha_{\text{cw}} = 4r$).
All $W^{(\ell)}$ remain \emph{frozen}; $A^{(\ell)}$, $B^{(\ell)}$, batch-normalization, and fully connected layers are trainable and included in the reported adapter size.
We initialize $A^{(\ell)}$ with Kaiming-uniform and set $B^{(\ell)}\!=\!0$, so each layer starts from $\tilde{W}^{(\ell)}\!=\!W^{(\ell)}$. Since CWLoRA is merged directly into the model weights, it introduces no additional inference-time overhead.

\section{Results}
\label{sec:results}

\begin{figure*}[t]
\centering
\setlength{\tabcolsep}{2pt}

\resizebox{1.0\textwidth}{!}{%
% 7 Columns total. 'm' columns ensure perfect vertical centering for images.
% Col 1: Inputs
% Col 2: Rotated Labels (with custom hskip to separate them visually)
% Col 3-7: The 5 Methods
\newcommand{\thead}[1]{\makebox[\linewidth][c]{\scriptsize\textbf{#1}}}
\begin{tabular}{ >{\centering\arraybackslash}m{0.14\textwidth} @{\hskip 16pt} c @{\hskip 10pt} *{5}{>{\centering\arraybackslash}m{0.14\textwidth}} }

% ==================== HEADER ROW ====================
& 
& \thead{TrustMarkP} 
& \thead{RoSteALS} 
& \thead{InvisMark} 
& \thead{HiDDeN} 
& \thead{WAM} \\[3pt]

% ==================== CAT IMAGES ====================

% --- Row 1: Target Images ---
\includegraphics[width=\linewidth]{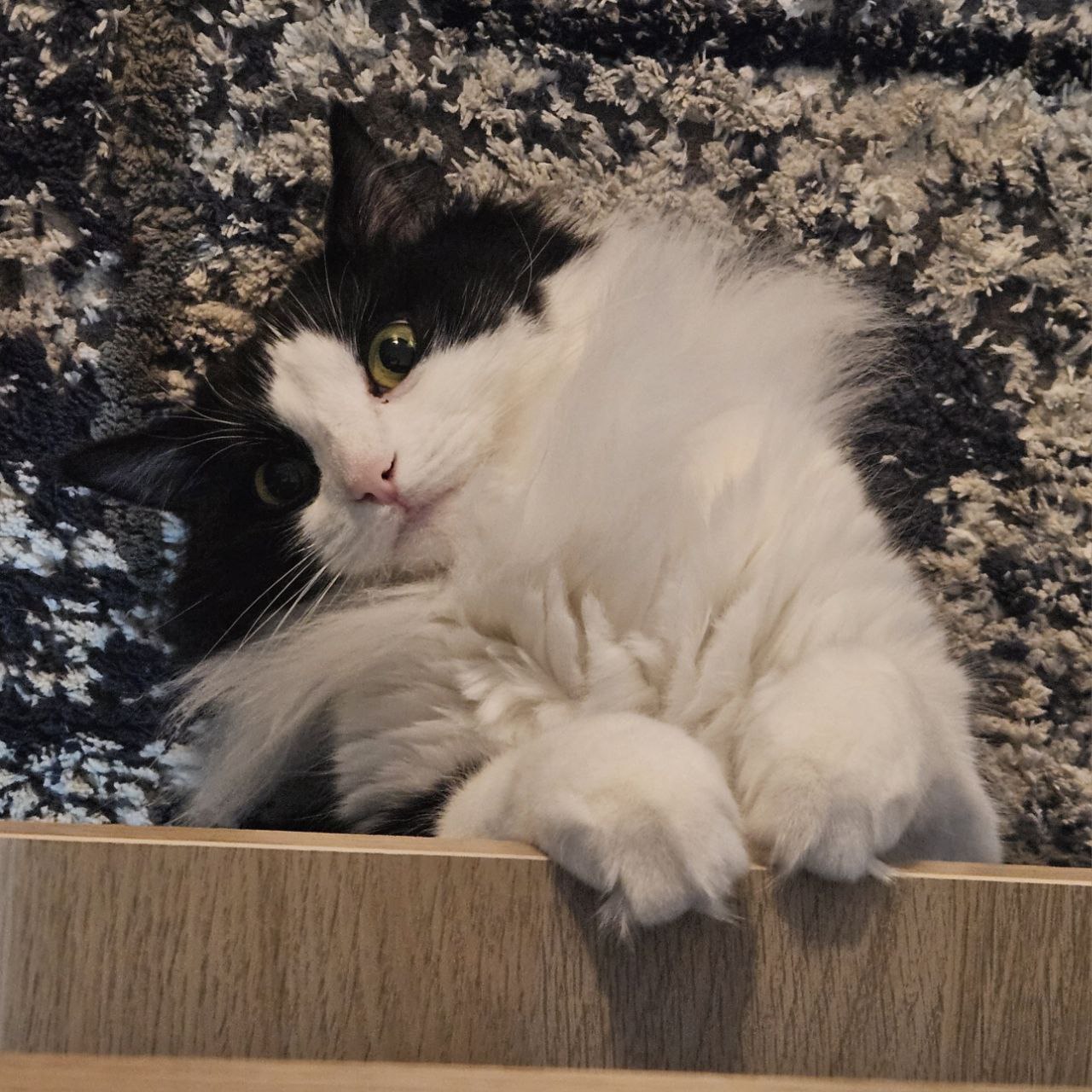} &
\raisebox{-0.4\height}{\rotatebox{90}{\textbf{Target}}} &
\includegraphics[width=\linewidth]{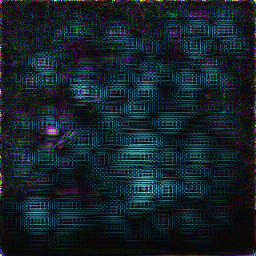} &
\includegraphics[width=\linewidth]{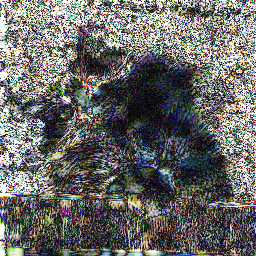} &
\includegraphics[width=\linewidth]{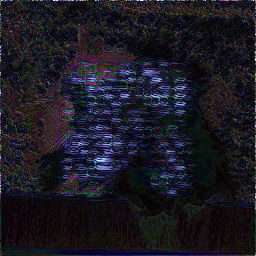} &
\includegraphics[width=\linewidth]{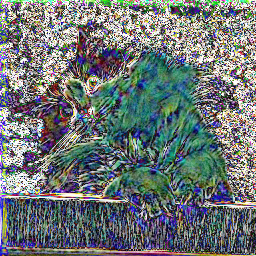} &
\includegraphics[width=\linewidth]{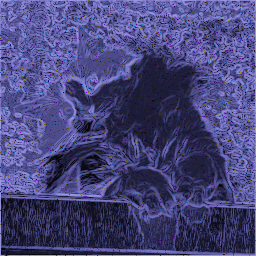} \\[-3pt]

% --- Row 2: Text below Col 1 only ---

{\scriptsize Original} & & & & & & \\[2pt]

% --- Row 3: Adapted Images ---
\includegraphics[width=\linewidth]{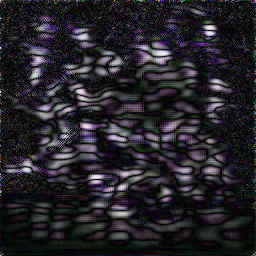} &
\raisebox{-0.45\height}{\rotatebox{90}{\textbf{Adapted}}} &
\includegraphics[width=\linewidth]{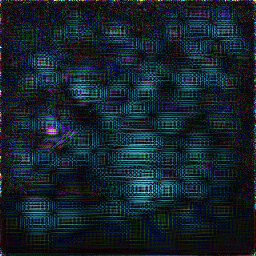} &
\includegraphics[width=\linewidth]{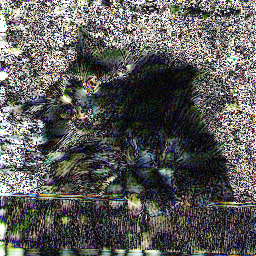} &
\includegraphics[width=\linewidth]{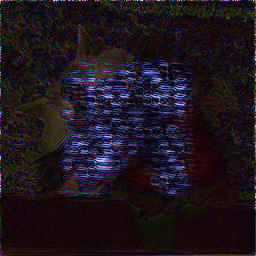} &
\includegraphics[width=\linewidth]{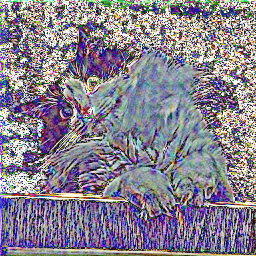} &
\includegraphics[width=\linewidth]{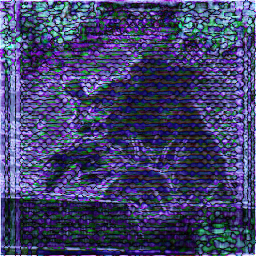} \\[-3pt]

% --- Row 4: Text below Col 1 only ---
\begin{tabular}{@{}c@{}}\scriptsize Base residual \\[-2pt] \scriptsize (TrustMarkQ)\end{tabular} & & & & & & \\

\end{tabular}
}% end resizebox

% \vspace{0.2cm}
\caption{\textbf{Visualization of residual alignment for encoder adaptation.}
The first column shows the input references: the original image and the base residual from TrustMarkQ. The remaining columns show the corresponding target and adapted residuals for each method.}
\label{fig:residual_grid}
\vspace{-0.2cm}
\end{figure*}

\begin{table}[t]
\centering
\caption{\textbf{Encoder--decoder adaptation across multiple base methods.} $\text{PSNR}_\text{tgt}$: imperceptibility of the target encoder ($\mathrm{PSNR}[\Etarget]$, dB). $\text{PSNR}_\text{ad}$: imperceptibility of adapted encoder ($\mathrm{PSNR}[\Eft]$, dB). $\text{ACC}_\text{tgt}$: noised bit accuracy of the target pipeline ($\mathrm{Acc}[\Etarget,\Dtarget]$). $\text{ACC}_\text{ad}$: noised bit accuracy of adapted pipeline ($\mathrm{Acc}[\Eft,\Dft]$). All accuracies are measured \emph{after} applying target's noise module. Clean accuracies consistently match the target across all settings (see Appendix).}
\label{tab:multi_base_encdec}
\renewcommand{\arraystretch}{1.15}
\setlength{\tabcolsep}{5pt}
\resizebox{\columnwidth}{!}{%
\begin{tabular}{l l c c c c}
\toprule
\textbf{Base} & \textbf{Target} & $\text{PSNR}_\text{tgt}$ & $\text{PSNR}_\text{ad}$ & $\text{ACC}_\text{tgt}$ & $\text{ACC}_\text{ad}$ \\
\midrule
\multirow{4}{*}{TrustMarkQ}
  & TrustMarkP  & 48.33 & 43.1  & 0.91 & 0.92 \\
  & RoSteALS    & 28.42 & 30.2  & 0.97 & 0.99 \\
  & InvisMark   & 47.37 & 44.2  & 0.93 & 0.98 \\
  & WAM         & 36.96 & 38.2  & 0.89 & 0.90 \\
\midrule
\multirow{4}{*}{TrustMarkP}
  & TrustMarkQ  & 39.69 & 38.85 & 0.95 & 0.96 \\
  & RoSteALS    & 28.42 & 29.78 & 0.97 & 0.99 \\
  & InvisMark   & 47.37 & 44.15 & 0.93 & 0.95 \\
  & WAM         & 36.96 & 38.15 & 0.89 & 0.91 \\
\midrule
\multirow{4}{*}{InvisMark}
  & TrustMarkQ  & 39.69 & 38.98 & 0.95  & 0.92  \\
  & TrustMarkP  & 48.33 & 41.37 & 0.91  & 0.89  \\
  & RoSteALS    & 28.42 & 30.66 & 0.97  & 0.98  \\
  & WAM         & 36.96 & 38.3  & 0.89 & 0.89 \\
\bottomrule
\end{tabular}%
}
\end{table}

\subsection{Experimental Setup}
\label{sec:exp_setup}

\vspace{0.2cm}
\noindent \textbf{Dataset and Augmentations.}
We fine-tune on a subset of MIRFLICKR~\cite{mirflickr} with 80k training and 10k test images; all results are reported on the test split.
When adapting to a target watermark, we use that method's \emph{reported} training noise set and strengths.
Our stochastic noise module, which is applied identically in training and evaluation, randomly samples a subset of these noises and composes them on the image.
Details of the module and the per-target noise sets/strengths are provided in the Appendix~\ref{sec:noise_modules}.

\vspace{0.2cm}
\noindent \textbf{Training Parameters.}
All fine-tuning runs use 150 epochs, a batch size of 128, and a learning rate of $4\times10^{-5}$. Training begins with a warm-up phase on a fixed batch (no augmentation) until bit accuracy reaches $0.90$,\footnote{$\mathrm{Acc}[E,D]$ denotes bit accuracy (fraction of recovered bits) when decoding messages embedded by encoder $E$ using decoder $D$. For encoder--decoder adaptation we monitor $\min(\mathrm{Acc}[\Eft,\Dtarget],\mathrm{Acc}[\Etarget,\Dft])$; for decoder-only we monitor $\mathrm{Acc}[\Etarget,\Dft]$.} after which we switch to the full training set. Once accuracy reaches $0.95$, the target noise module $\mathcal{A}_\text{tgt}$ is linearly ramped to full strength over $\approx$16 epochs. For encoder--decoder adaptation, the reconstruction-loss weights $\alpha,\beta,\gamma,\delta$ (Eq.~\ref{eq:rec-loss}) are set to 1, and the secret-loss weights $\mu_2$ (Eq.~\ref{eq:enc-loss-definition}) and $\lambda_1$ (Eq.~\ref{eq:dec-paired}) are ramped from 0 to 1 over 10k iterations. Adaptation time is comparable to standard training (Appendix~\ref{sec:time_analysis}).

\vspace{0.2cm}
\noindent \textbf{Watermarking Methods.}
We provide experiments on TrustMark (P and Q)~\cite{bui2023trustmark}, RoSteALS~\cite{bui2023rosteals}, InvisMark~\cite{xu2025invismark}, HiDDeN~\cite{zhu2018hidden}, and WAM~\cite{sander2024watermark}. We use the public official checkpoints for all methods\footnote{Except for HiDDeN for which we use the implementation from \href{https://github.com/ando-khachatryan/HiDDeN}{https://github.com/ando-khachatryan/HiDDeN}}.
The secret key lengths are 30 for HiDDeN, 32 for WAM, and 100 for other methods.
The encoder/decoder architecture types and parameter counts are reported in Appendix~\ref{sec:config_details} (Table~\ref{tab:wm_params}).

\vspace{0.2cm}
\noindent \textbf{Model Adapters.}
We consider convolutional adapters including IA$^3$ \cite{ia3}, SE-Gate \cite{se-gate}, $1{\times}1$-Bottleneck \cite{1x1-bn}, PEFT-LoRA \cite{peft}, and CWLoRA (Section~\ref{sec:method_cwlora}). Details on the adapters and their implementations are provided in Appendix~\ref{sec:adapter_details}.
Fig.~\ref{fig:adapter_comparison} compares these adapters; we adopt CWLoRA-16 throughout (see Section~\ref{sec:exp_deconly}). For InvisMark's ConvNeXt decoder, a variant is used (Appendix~\ref{sec:adapter_details}).

\subsection{Decoder-only Adaptation}
\label{sec:exp_deconly}

\noindent \textbf{Which adapter is most effective?}
Fig.~\ref{fig:adapter_comparison} presents the performance of adapted decoders (bit accuracy under the target noise module) using different adapter types. The adaptation is performed with TrustMarkQ as the base method and three target methods (i.e., InvisMark, WAM, TrustMarkP). The figure also includes the original target decoder \(\Dtarget\) as a baseline. The \emph{Full} label denotes fine-tuning \emph{all} parameters of the base decoder.
As shown, some adapters approach \(\Dtarget\)'s performance with far fewer trainable parameters. Among them, CWLoRA consistently provides a strong accuracy–size trade-off, especially at rank 16, and outperforms common post-layer adapters, indicating that direct weight-space updates are better matched to convolutional watermarking backbones. Adapters that fail in Fig.~\ref{fig:adapter_comparison}, including IA$^3$, also train batch-normalization and the final linear layer, so those layers alone do not account for the adaptation. Accordingly, we use CWLoRA-16 in subsequent decoder-only and encoder–decoder experiments. We further observe in Figure~\ref{fig:adapter_comparison} that lightweight adapters (e.g., CWLoRA-16/32) can outperform full fine-tuning of a weaker base-decoder configuration (i.e., ResNet-18 TrustMarkQ decoder as $\Dbase$), suggesting that adapted decoders effectively use the frozen features of \(\Dbase\) to decode messages from \(\Etarget\).

\vspace{0.2cm}
\noindent \textbf{Adaptation across base decoders.}
Next, we evaluate decoder-only adaptation across multiple base decoders \(\Dbase\).
Table~\ref{tab:base-decoder-comparison} summarizes results of adapting several bases to multiple targets.
In most cases, the lightweight CWLoRA-16 adapter attains clean/noised bit accuracy close to the \(\mathrm{Acc}[\Etarget,\Dtarget]\) baseline.
We also include an ImageNet-pretrained ResNet-50 as $D_{base}$. While it succeeds for some targets, training fails (remaining at the 0.5 random-guessing baseline) for highly imperceptible targets like InvisMark and TrustMarkP (Table~\ref{tab:base-decoder-comparison}). We hypothesize that without pre-learned watermarking priors (e.g., frequency sensitivity), the optimization landscape for these highly imperceptible signals is too difficult for a lightweight adapter to traverse from scratch.

% For some targets, specific bases achieve slightly better results.
% For example, with WAM as the target, using InvisMark as the base yields $0.925$ noised bit accuracy versus $0.905$ with TrustMarkQ; conversely, with RoSteALS as the target, TrustMarkQ reaches $0.965$ while InvisMark attains $0.959$.
% These patterns suggest method-dependent compatibility, likely influenced by architectural similarity and robustness to particular noise types.

\vspace{0.2cm}
\noindent \textbf{What governs base-target compatibility?}
To further analyze the underlying factors governing these base-target compatibilities, we provide a representational similarity study using Centered Kernel Alignment (CKA) between decoder pairs in Appendix~\ref{sec:cka_analysis}. Our findings indicate that high feature-space overlap alone does not guarantee successful transfer; adaptability depends on target imperceptibility, watermark-pattern similarity, noise robustness, and base architectural capacity.

\begin{table}[t]
\centering
\caption{\textbf{CWLoRA on a strong base vs.\ full fine-tuning of a compact model.} Both approaches use a comparable number of trainable parameters, but CWLoRA on TrustMarkQ preserves the base's perceptual quality while full fine-tuning of TrustMarkQ-Small suffers a large PSNR drop at similar accuracy.}
\label{tab:base_necessity}
\renewcommand{\arraystretch}{1.15}
\setlength{\tabcolsep}{4pt}
\resizebox{\columnwidth}{!}{%
\begin{tabular}{llccc}
\toprule
\textbf{Target} & \textbf{Method} & $\mathrm{PSNR}[\Eft]$ & $\mathrm{Acc}[\Eft,\Dft]$ & $\mathrm{Acc}[\Etarget,\Dtarget]$ \\
\midrule
\multirow{2}{*}{InvisMark}
  & TrustMarkQ-Small (Full) & 35.17 & 0.97 & 0.93 \\
  & \textbf{TrustMarkQ (CWLoRA)} & \textbf{44.20} & \textbf{0.98} & 0.93 \\
\midrule
\multirow{2}{*}{TrustMarkP}
  & TrustMarkQ-Small (Full) & 37.07 & 0.91 & 0.91 \\
  & \textbf{TrustMarkQ (CWLoRA)} & \textbf{43.10} & \textbf{0.92} & 0.91 \\
\bottomrule
\end{tabular}%
}
\end{table}

\begin{table}[t]
\centering
\caption{\textbf{OOD and adversarial robustness.} Each cell reports \emph{Adapted (Ours) / Target (Original)} accuracy after the listed perturbation. Arrows indicate the direction of higher robustness. Adapted models broadly inherit the target's OOD invariances.}
\label{tab:ood_robustness}
\renewcommand{\arraystretch}{1.15}
\setlength{\tabcolsep}{3.5pt}
\resizebox{\columnwidth}{!}{%
\begin{tabular}{lccccc}
\toprule
\textbf{Target} & \textbf{Geometric} $\uparrow$ & \textbf{Valuemetric} $\uparrow$ & \textbf{Regeneration} $\uparrow$ & \textbf{Spoofing} $\downarrow$ & \textbf{BBAdv} $\uparrow$\\
\midrule
InvisMark & 0.57 / 0.69 & 0.65 / 0.65 & 0.50 / 0.50 & 0.72 / 0.81 & 0.95 / 0.91 \\
RoSteALS  & 0.52 / 0.52 & 0.97 / 0.95 & 0.89 / 0.79 & 0.99 / 0.96 & 1.00 / 0.97 \\
TrustMarkP   & 0.70 / 0.68 & 0.68 / 0.69 & 0.51 / 0.51 & 0.85 / 0.76 & 0.63 / 0.56 \\
WAM      & 0.84 / 0.83 & 0.68 / 0.75 & 0.51 / 0.50 & 0.89 / 0.96 & 0.94 / 0.98 \\
\bottomrule
\end{tabular}%
}
\end{table}

\begin{table}[t]
\centering
\caption{\textbf{Cross-dataset evaluation on AI-generated images.} Adapted models (TrustMarkQ base) evaluated on the Defactify Image Dataset, outside the MIRFLICKR training distribution. Metrics remain stable, confirming that adaptation does not overfit to the training dataset.}
\label{tab:ai_gen}
\renewcommand{\arraystretch}{1.15}
\resizebox{\columnwidth}{!}{%
\begin{tabular}{lccc}
\toprule
\textbf{Target} & $\mathrm{PSNR}[\Eft]$ & $\mathrm{Acc}[\Eft,\Dft]$ & $\mathrm{Acc}[\Etarget,\Dtarget]$ \\
\midrule
InvisMark & 44.12 & 0.98 & 0.93 \\
RoSteALS  & 30.26 & 0.98 & 0.97 \\
TrustMarkP    & 42.83 & 0.92 & 0.91 \\
WAM       & 38.15 & 0.91 & 0.89 \\
\bottomrule
\end{tabular}%
}
\end{table}

\subsection{Encoder-Decoder Adaptation}
\label{sec:exp_encdec}

The encoder–decoder adaptation setting and objectives were introduced in Section~\ref{sec:method_encdec}.
In our experiments, we adapt from multiple base methods to several target methods.
Fine-tuning follows Eq.~\ref{eq:enc-loss-definition} for the encoder and Eq.~\ref{eq:dec-paired} for the decoder, using the CWLoRA-16 adapter.

\vspace{0.2cm}
\noindent \textbf{Does adaptation generalize across base methods?}
Table~\ref{tab:multi_base_encdec} summarizes encoder--decoder adaptation results across three base methods (TrustMarkQ, TrustMarkP, InvisMark) and four target methods each.
% For each base--target pair, we report the target's imperceptibility ($\mathrm{PSNR}[\Etarget]$), our adapted encoder's imperceptibility ($\mathrm{PSNR}[\Eft]$), and end-to-end noised bit accuracy ($\mathrm{Acc}[\Eft,\Dft]$ vs.\ the target baseline $\mathrm{Acc}[\Etarget,\Dtarget]$).
Across all bases, $\mathrm{PSNR}[\Eft]$ tracks $\mathrm{PSNR}[\Etarget]$ within a few dB for most pairs, with the largest gaps occurring for the highest-fidelity targets (TrustMarkP at 48.3\,dB and InvisMark at 47.4\,dB); the adapted encoder nonetheless stays above 29\,dB in every case, confirming that imperceptibility is largely preserved regardless of the choice of base.
% Residual visualizations, SSIM, and LPIPS are in Appendices~\ref{sec:adaptation_visualizations} and~\ref{sec:extra_qual_metrics}.
See Figure~\ref{fig:residual_grid} and Appendix~\ref{sec:adaptation_visualizations} for residual visualizations. SSIM and LPIPS metrics are in Appendix~\ref{sec:extra_qual_metrics}.
The noised accuracy $\mathrm{Acc}[\Eft,\Dft]$ remains competitive with the target baseline across most pairs, demonstrating that adaptation generalizes beyond a single base and that the choice of base method does not critically constrain the framework's effectiveness.
A comprehensive single-base analysis with per-target clean/noised breakdowns across four metric panels, including $\mathrm{Acc}[\Etarget,\Dft]$ and $\mathrm{Acc}[\Eft,\Dtarget]$, is provided in Appendix~\ref{sec:encdec_appendix}.

\begin{table}[t]
    \centering
    \caption{\textbf{Comparison of approaches for multi-decoder inference.} The Oracle rows are the upper bound for each decoder set. Acc is full-message accuracy.}
    \label{tab:multi-dec}
    \renewcommand{\arraystretch}{1.15}
    \setlength{\tabcolsep}{3pt}
    \resizebox{\columnwidth}{!}{%
    \begin{tabular}{lcccc}
    \toprule
    \multirow{2}{*}{Method} & \multicolumn{2}{c}{Acc (\%)} & \multirow{2}{*}{Size (GB)} & \multirow{2}{*}{Time (ms)} \\
    \cmidrule(lr){2-3}
    & Clean & Noised & & \\
    \midrule
    Original+Oracle      & 96.70 & 86.17 & 1.19 & 1.35 \\
    Adapter+Oracle       & 97.23 & 86.19 & 0.14 & 0.62 \\
    \midrule
    Original+Classifier  & 96.30 & 85.66 & 1.29 & 1.79 \\
    Adapter+Classifier   & \textbf{96.87} & \textbf{85.73} & 0.24 & \textbf{1.07} \\
    Original+ECC         & 96.13 & 85.68 & 1.19 & 4.75 \\
    Adapter+ECC          & 95.55 & 84.66 & \textbf{0.14} & 2.18 \\
    \bottomrule
    \end{tabular}%
    }
    \end{table}

% \vspace{0.2cm}
% \noindent \textbf{Adapted encoders preserve imperceptibility.}
% The adapted encoders produce residuals that visually align with \(\Etarget\) while shifting significantly from \(\Ebase\), with only lightweight updates (visualizations in Appendix~\ref{sec:adaptation_visualizations}). SSIM and LPIPS evaluations confirm that perceptual quality is preserved (Appendix~\ref{sec:extra_qual_metrics}).

\vspace{0.2cm}
\noindent \textbf{Adapted decoders preserve robustness.}
Despite substantial architectural mismatches (e.g., RoSteALS's VQGAN or WAM's ViT), adaptation yields decoders that recover messages under the target's noise module with similar accuracies, even when the base has not been trained on those degradations, indicating transfer of the target's invariances.

\vspace{0.2cm}
\noindent \textbf{Cross-compatibility between adapted and original models.}
The adapted models remain \emph{interoperable} with the original target models: the adapted encoder's residuals are decodable by the original target decoder ($\mathrm{Acc}[\Eft,\Dtarget]$), and the adapted decoder can recover messages from the original target encoder ($\mathrm{Acc}[\Etarget,\Dft]$). Both cross-accuracies closely track the target baseline (Appendix~\ref{sec:encdec_appendix}, Figure~\ref{fig:e2e_performance}), confirming that adaptation captures the functionally relevant behavior of the target rather than merely learning a self-consistent pair. This holds even when residual alignment is imperfect.

\vspace{0.2cm}
\noindent \textbf{CWLoRA achieves efficiency beyond parameter count.}
We compare against full fine-tuning of a compact ``TrustMarkQ-Small'' model with a comparable trainable-parameter budget (Table~\ref{tab:base_necessity}). At similar robustness, full fine-tuning of the smaller model suffers a substantially larger PSNR drop than CWLoRA on the strong base, confirming that the base's pre-existing priors, not the parameter count alone, drive the benefit.

\vspace{0.2cm}
\noindent \textbf{Robustness beyond training noise.}
Table~\ref{tab:ood_robustness} evaluates four target methods under five out-of-distribution perturbation categories that are strictly stronger than, or entirely absent from, the training profiles: \textbf{Geometric} perturbations (stronger crops, rotations, and perspective distortions), \textbf{Valuemetric} perturbations (stronger brightness, contrast, and color jitter), \textbf{Regeneration} via Stable Diffusion v1.5 image purification~\cite{an2024waves,saberi2024robustness}, adversarial \textbf{Spoofing} attacks~\cite{saberi2024robustness}, and a \textbf{black-box adversarial attack} (BBAdv)~\cite{hu2025transfer,mueller2025blackbox} using TrustMarkQ as the surrogate decoder.
The table reports adapted vs.\ original-target accuracy. The adapted models broadly inherit the target's OOD invariances, with gaps usually small and occasionally in the adapted model's favor (e.g., RoSteALS under Regeneration: 0.89 vs.\ 0.79). Under BBAdv, the adapted models show no systematic degradation despite the structural disadvantage, and in some cases are more robust. While this does not establish a guarantee, it empirically shows that the base's adversarial weaknesses do not leak into adapted models on these attacks.

\vspace{0.2cm}
\noindent \textbf{Generalization across datasets.}
To verify that the adapted models do not overfit to MIRFLICKR~\cite{mirflickr}, we evaluate on AI-generated images from the Defactify Image Dataset~\cite{an2024waves}. Table~\ref{tab:ai_gen} reports imperceptibility and noised accuracy for four targets using TrustMarkQ as the base. The metrics remain stable compared to in-distribution results (Table~\ref{tab:multi_base_encdec}), confirming that the adaptation learns transferable watermark transformations rather than dataset-specific artifacts.

\vspace{0.2cm}
\noindent \textbf{Effect of the residual-matching loss.}
Even when the encoder-side secret loss is \emph{omitted} ($\mu_2{=}0$), $\mathrm{Acc}[\Eft,\Dtarget]$ can improve purely from the residual-matching objective (Appendix~\ref{sec:res_only_adapt}). An ablation on the loss terms (Appendix~\ref{sec:res_hparam_analysis}) confirms that all four components together yield the strongest residual alignment.

\subsection{Multi-Decoder Inference}
\label{sec:multi-dec}

We analyze a practical application of the adapterized framework: decoding messages from images watermarked by \emph{multiple} methods.
Given a query image (watermarked by any method in the pool), the goal is to recover the full message.
Accordingly, we report full-message accuracy (percentage of images whose data bits are fully correct after ECC), not bit accuracy.
For ECC, we apply BCH error-correction~\cite{bch}: for 100-bit watermarks we use 40 data + 56 ECC bits; for WAM (32 bits) we use 16 data + 16 ECC bits.

\vspace{0.2cm}
\noindent \textbf{Model sets.}
To support multi-decoder inference, one could load each method's original decoder (\textit{Original}) or, instead, load a single base decoder plus lightweight adapters (\textit{Adapter}).
We consider both settings using the methods \{TrustMarkQ, TrustMarkQ\_v2, TrustMarkQ\_low, TrustMarkP, InvisMark, RoSteALS, WAM\}.
TrustMarkQ\_low is trained with milder noise (same noise setting as TrustMarkP); TrustMarkQ\_v2 differs only by initialization seed and serves to test closely related watermark residuals.
Despite their similarity, cross-decoding between TrustMarkQ and TrustMarkQ\_v2 is near random (bit accuracy $\approx 0.5$), underscoring the need for decoder selection.

With fp32 weights, moving from \textit{Original} to \textit{Adapter} reduces total decoder size from $\sim$1.2\,GB to $\sim$0.14\,GB (Table~\ref{tab:multi-dec}). This 8$\times$ memory reduction demonstrates advantages of adaptation: rather than deploying an ensemble of full targets, the system ships a single base and compact deltas.
% The 1.19\,GB Oracle and ECC rows are the seven full decoders, with no base encoder and no stored ECC network; the 0.14\,GB adapter rows are one TrustMarkQ decoder plus six CWLoRA decoders. The classifier rows add the ResNet-50 selector (0.10\,GB).

\vspace{0.2cm}
\noindent \textbf{Decoder selection.}
At test time, we select the right decoder via: (i) \textbf{ECC-based}, run decoders in random order and pick the first whose BCH verification passes; (ii) \textbf{Classifier-based}, a ResNet-50 (Appendix~\ref{sec:selection_classifier_train}) predicts the watermark method, then only the predicted decoder is used; (iii) \textbf{Oracle}, assumes the true label is known (upper bound). Pseudo-codes are in Appendix~\ref{sec:multi_dec_inf_pseudo}. The ECC approach scales $O(N)$ with pool size; the classifier is $O(1)$ but requires retraining when new watermarks are added.

\vspace{0.2cm}
\noindent \textbf{Results.}
Table~\ref{tab:multi-dec} reports full-message accuracy, size, and runtime; exact error rates are in Appendix Table~\ref{tab:multi-dec-errors}.
Using adapters, both \textit{Adapter+ECC} and \textit{Adapter+Classifier} achieve accuracy close to the corresponding original-decoder baselines while reducing loaded model size by $\sim\!8\times$.
Moreover, \textit{Adapter+\allowbreak Oracle} closely matches \textit{Original+Oracle}, indicating that adapterization itself introduces minimal loss.
Overall, these results demonstrate a lightweight and extensible multi-watermark pipeline: new methods can be added by shipping an additional adapter, without deploying full decoder models. Detailed per-method results are provided in the Appendix~\ref{sec:multi_dec_inf_perwm}.

\section{Conclusion}

We presented a parameter-efficient framework for transforming neural watermarking models using CWLoRA adapters trained via multi-domain residual alignment and target-noise secret losses. Our adapted models match target residuals and decoding robustness across diverse architectures, generalize to multiple base methods, out-of-distribution perturbations, and unseen datasets, and our multi-decoder pipeline preserves accuracy while reducing the loaded footprint by $\sim8\times$, enabling a unified, interoperable framework that can switch between watermarking styles without retraining full models.

Future work includes black-box adaptation when the target is available only as an API or binary, extending to in-generation and autoregressive watermarks, native ViT adapters for non-CNN bases, and shared-parameter designs like VB-LoRA~\cite{vblora}.
{
    \small
    \bibliographystyle{ieeenat_fullname}
    \bibliography{main}
}

\clearpage
\appendix
\maketitlesupplementary

% ====================================================================
% APPENDIX A: Comprehensive Encoder-Decoder Analysis
% ====================================================================
\section{Comprehensive Encoder-Decoder Adaptation Analysis}
\label{sec:encdec_appendix}

% \begin{figure*}[t]
%     \centering
%     \includegraphics[width=0.98\linewidth]{images/wm_adapt_2x2_v2.pdf}
%     \caption{Caption}
%     \label{fig:encoder_results}
% \end{figure*}

\begin{figure*}[t]
    \centering
    \includegraphics[width=0.98\linewidth]{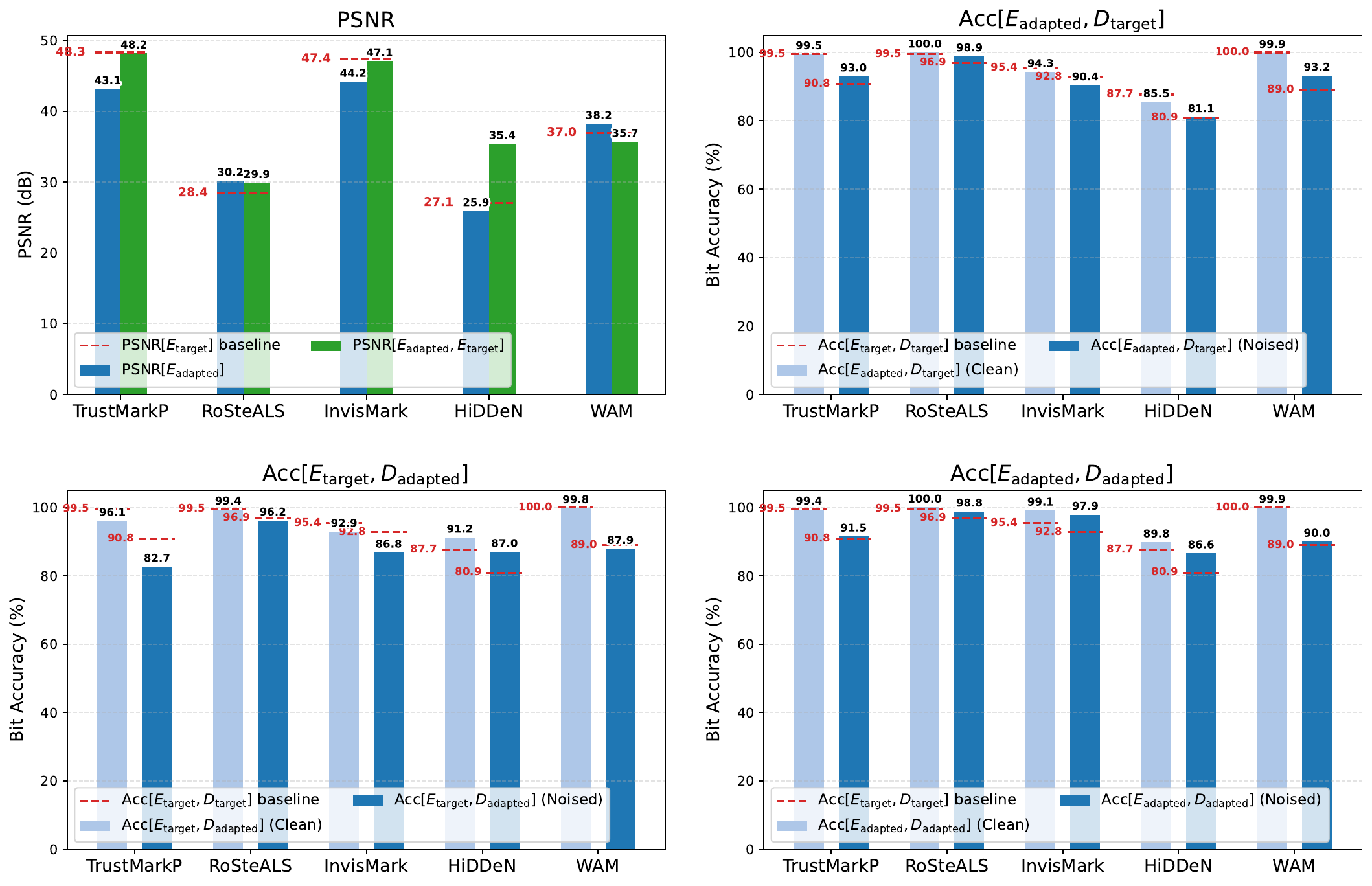}
    \caption{\textbf{Encoder-decoder adaptation performance.} The base method (\(\Ebase,\Dbase\)) is TrustMarkQ. \textbf{Top-left:} \(\mathrm{PSNR}\) between the input and output of $\Eft$ (\(\mathrm{PSNR}[\Eft]\), blue), and between outputs of $\Eft$ and $\Etarget$ (\(\mathrm{PSNR}[\Eft,\Etarget]\), green); the red dashed line is \(\mathrm{PSNR}[\Etarget]\) (between input and output of $\Etarget$). \textbf{Top-right:} \(\mathrm{Acc}[\Eft,\Dtarget]\) (clean/noised) with red dashed baseline \(\mathrm{Acc}[\Etarget,\Dtarget]\). \textbf{Bottom-left:} \(\mathrm{Acc}[\Etarget,\Dft]\) (clean/noised). \textbf{Bottom-right:} \(\mathrm{Acc}[\Eft,\Dft]\) (clean/noised). All ``noised'' metrics use the target method's noise set for training and evaluation.}

    \label{fig:e2e_performance}
\end{figure*}

Figure~\ref{fig:e2e_performance} provides a detailed analysis of encoder--decoder adaptation using TrustMarkQ as the base, with four panels: (top-left) $\mathrm{PSNR}[\Eft]$ and $\mathrm{PSNR}[\Eft,\Etarget]$ against the $\mathrm{PSNR}[\Etarget]$ baseline; (top-right) $\mathrm{Acc}[\Eft,\Dtarget]$ (clean/noised) against the target baseline; (bottom-left) $\mathrm{Acc}[\Etarget,\Dft]$ (clean/noised); and (bottom-right) $\mathrm{Acc}[\Eft,\Dft]$ (clean/noised). Clean accuracy consistently matches the target across all settings.

% ====================================================================
% APPENDIX B: Adaptation Visualizations and Ablations
% ====================================================================
\section{Adaptation Visualizations and Ablations}

\subsection{Residual-Matching-only Adaptation}
\label{sec:res_only_adapt}

\begin{figure*}
    \centering
    \includegraphics[width=0.98\linewidth]{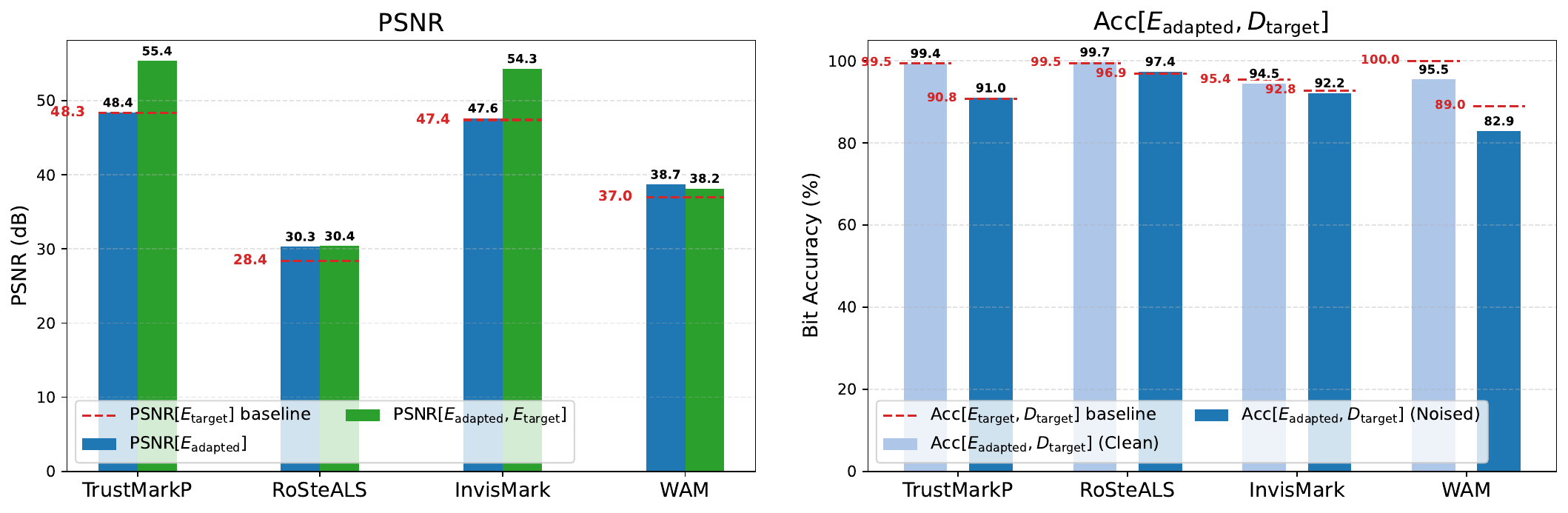}
    \caption{\textbf{Performance of residual-matching-only encoder adaptation.} Setting $\mu_2=0$ results in higher PSNR values compared to Figure~\ref{fig:e2e_performance}, while surprisingly, maintaining high decodability by $\Dtarget$.}

    \label{fig:enc_nostw_performance}
\end{figure*}

Figure~\ref{fig:enc_nostw_performance} reports encoder adaptation results (TrustMarkQ as the base) when training \emph{only} with the reconstruction loss, i.e., setting $\mu_2{=}0$ in Eq.~\ref{eq:enc-loss-definition} and optimizing $\mathcal{L}_\text{rec}$.
With fewer training constraints, the adapted encoders achieve higher \(\mathrm{PSNR}[\Eft,\Etarget]\) and \(\mathrm{PSNR}[\Eft]\) than in Figure~\ref{fig:e2e_performance}.
Perhaps surprisingly, even without explicitly optimizing decodability, \(\mathrm{Acc}[\Eft,\Dtarget]\) remains competitive, indicating that strong residual alignment alone can yield messages recoverable by the target decoder.

\subsection{Residual-Matching Loss Analysis}
\label{sec:res_hparam_analysis}

\begin{table*}[t]
\centering
\caption{\textbf{The effect of residual-matching loss terms.} Based on the results, using all four loss terms from Equation~\ref{eq:rec-loss} achieves the best overall performance.
% \textbf{Effect of loss-term weighting on encoder adaptation.} Each row reports PSNR scores for TrustMarkP and WAM under different combinations of loss coefficients \(\alpha,\beta,\gamma,\delta\).
}
\label{tab:recon-params}
\renewcommand{\arraystretch}{1.15}
\resizebox{\textwidth}{!}{%
\begin{tabular}{l c c c c}
\toprule
\multirow{2}{*}{Loss Coefficients} &
\multicolumn{2}{c}{TrustMarkP} &
\multicolumn{2}{c}{WAM} \\
\cmidrule(lr){2-3}\cmidrule(lr){4-5}
& \shead{$\mathrm{PSNR}[E_{\text{adapted}}]$} 
& \shead{$\mathrm{PSNR}[E_{\text{adapted}},E_{\text{target}}]$}
& \shead{$\mathrm{PSNR}[E_{\text{adapted}}]$} 
& \shead{$\mathrm{PSNR}[E_{\text{adapted}},E_{\text{target}}]$} \\
\midrule
$\alpha = 1$ 
    & 46.0 & 47.7 & 38.1 & 36.0 \\

$\alpha = 1,\ \beta = 0.5$ 
    & \textbf{46.5} & 52.5 & \textbf{41.6} & 36.8 \\

$\alpha = 1,\ \gamma = 0.5$
    & 46.0 & 51.0 & 37.9 & 36.5 \\

$\alpha = 1,\ \delta = 0.5$
    & 45.1 & 53.2 & 38.3 & 37.8 \\

$\alpha = 1,\ \beta = 0.5,\ \gamma = 0.5,\ \delta = 0.5$
    & 45.9 & 54.1 & 38.8 & 38.3 \\

$\alpha = 1,\ \beta = 1,\ \gamma = 1,\ \delta = 1$
    & 46.1 & \textbf{54.8} & 39.0 & \textbf{38.4} \\
\bottomrule
\end{tabular}%
}
\end{table*}

Table~\ref{tab:recon-params} ablates the residual-matching objective in Eq.~\ref{eq:rec-loss} for encoder adaptation with TrustMarkQ as the base. To isolate the effect of the loss terms, we disable the noise module during training (unlike Fig.~\ref{fig:e2e_performance}). The ablation varies the weights \(\alpha,\beta,\gamma,\delta\) for the pixel, gradient, frequency, and cosine-similarity terms, respectively. Using all four terms together yields the strongest residual alignment, improving similarity to \(\Etarget\).

% ====================================================================
% APPENDIX C: Additional Analyses
% ====================================================================
\section{Additional Analyses}

\subsection{Perceptual Quality Analysis}
\label{sec:extra_qual_metrics}

To further evaluate the visual imperceptibility of the watermarks produced by our adapted encoders, we report the Structural Similarity Index Measure (SSIM) and Learned Perceptual Image Patch Similarity (LPIPS). The LPIPS metric is calculated with respect to an AlexNet backbone trained on ImageNet. As shown in Table \ref{tab:ssim_lpips}, the adapted encoders achieve perceptual quality scores that closely match those of the original target encoders across all evaluated methods. This demonstrates that our lightweight adaptation framework effectively preserves the delicate, high-frequency steganographic artifacts required for decoding without degrading the visual quality of the cover image beyond the target method's original footprint.

\begin{table*}[t]
    \centering
    \caption{Perceptual quality metrics (SSIM and LPIPS) for original target encoders and our adapted encoders. The base encoder used for adaptation is TrustMarkQ. LPIPS is calculated using an AlexNet backbone pre-trained on ImageNet.}
    \label{tab:ssim_lpips}
    
    \setlength{\tabcolsep}{10pt}
    \resizebox{\textwidth}{!}{
    \begin{tabular}{@{}lccccc@{}}
        \toprule
        \textbf{SSIM ($\uparrow$) / LPIPS ($\downarrow$)} & \textbf{TrustMarkP} & \textbf{RoSteALS} & \textbf{InvisMark} & \textbf{HiDDeN} & \textbf{WAM} \\
        \midrule
        \textbf{Target (Original)} & 1.00 / 0.00 & 0.84 / 0.04 & 1.00 / 0.00 & 0.81 / 0.22 & 0.96 / 0.03 \\
        \textbf{Adapted (Ours)} & 0.99 / 0.00 & 0.89 / 0.11 & 0.99 / 0.01 & 0.77 / 0.22 & 0.97 / 0.02 \\
        \bottomrule
    \end{tabular}
    }
\end{table*}

\subsection{Feature Similarity Analysis}
\label{sec:cka_analysis}

To understand why certain base-target pairs exhibit better compatibility during adaptation, we analyze the representational similarity between the features produced by the frozen base decoder and the target decoder. We utilize Centered Kernel Alignment (CKA) \cite{kornblith2019similarity}, a robust metric for quantifying likeness between deep neural network representations.

Because watermarking models can vary significantly in depth and architecture, we summarize the layer-by-layer CKA matrices into two network-level metrics: \textit{Peak CKA Similarity} (the maximum CKA score across all possible layer pairs between the two models) and \textit{Average Best-Match} (the average similarity when each layer in the base model is paired with its most similar layer in the target model).

\begin{figure*}[t]
\centering
\includegraphics[width=1.0\linewidth]{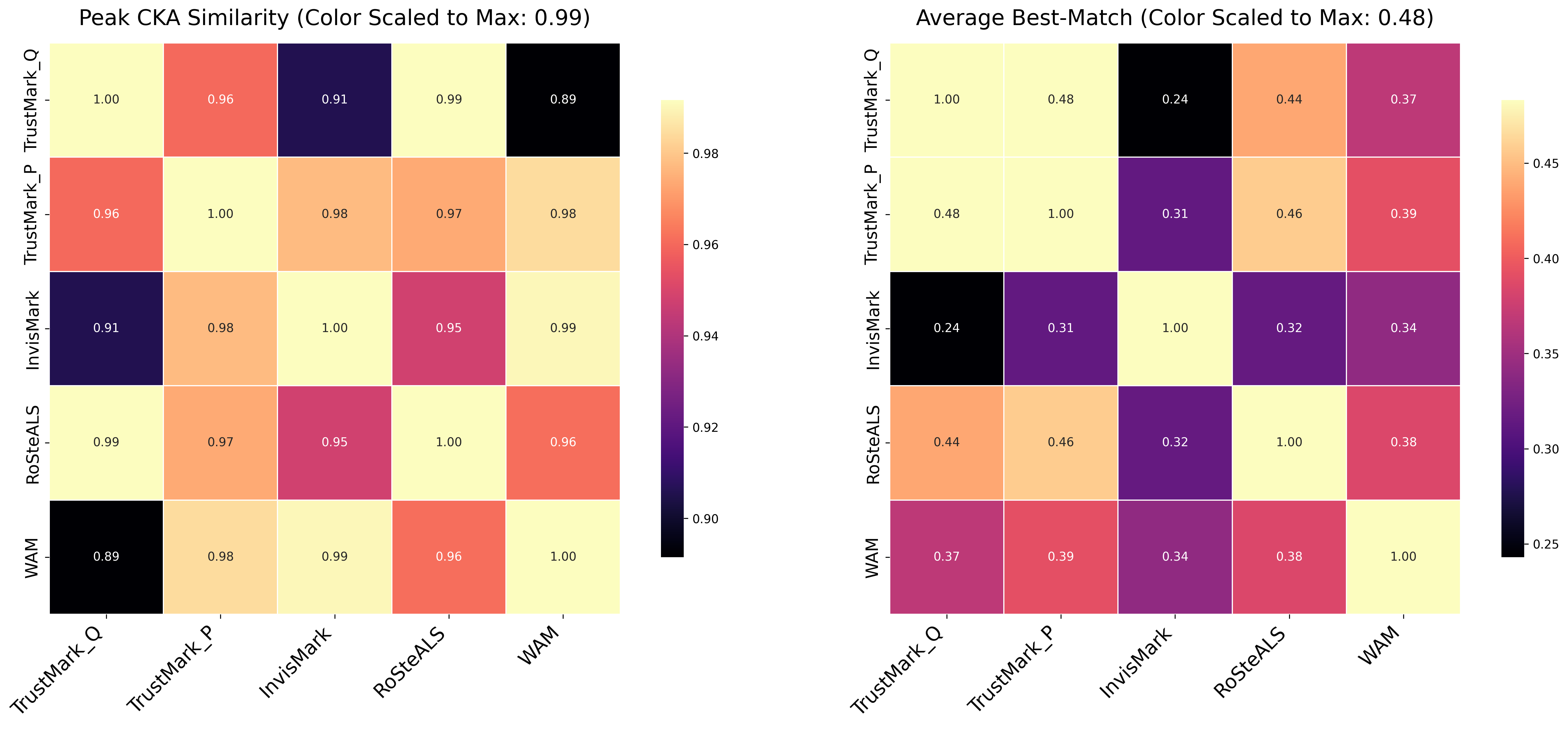} 
\caption{\textbf{Feature Similarity between base and target decoders.} Left: Peak CKA Similarity. Right: Average Best-Match CKA. TrustMarkQ and InvisMark exhibit the lowest similarity scores for InvisMark (0.91 Peak, 0.24 Average), yet TrustMarkQ serves as the best base decoder for adapting to InvisMark (Table~\ref{tab:base-decoder-comparison}). This demonstrates that strong representational overlap is not the sole factor for effective watermark transformation.}
\label{fig:cka_heatmap}
\end{figure*}

Fig.~\ref{fig:cka_heatmap} presents these metrics across several base-target decoder pairings. The analysis reveals that representational similarity does not reliably correlate with adaptation performance. For example, TrustMarkQ and InvisMark have a Peak CKA similarity of 0.91 and an Average Best-Match of 0.24. These are the lowest similarity scores observed between InvisMark and the other evaluated methods. However, as shown in Table~\ref{tab:base-decoder-comparison}, using TrustMarkQ as the base decoder achieves the highest adaptation accuracy to InvisMark, reaching 0.959 (clean) and 0.909 (noised).

This suggests that parameter-efficient adaptation via CWLoRA relies more on a complex interplay of architectural capacity and noise robustness rather than requiring explicit, pre-existing feature-space overlap. Consequently, the framework can successfully navigate distributional shifts even between decoders with lower representational similarity.

\subsection{Adaptation Training Time}
\label{sec:time_analysis}

Table~\ref{tab:runtime} details the training runtime across our various adaptation configurations, all of which use TrustMarkQ as the base method. All training was conducted using a single A100 (80GB) GPU. As noted in Section~\ref{sec:exp_setup}, all fine-tuning adaptations are run for 150 epochs to ensure convergence. For baseline context, training the original TrustMarkQ base model from scratch required 59.6 hours over 250 epochs.

In the decoder-only setups, we observe that training with the lightweight CWLoRA-16 adapter yields runtimes (25.4--30.6 hours) that are virtually identical to full-model fine-tuning (25.8--30.8 hours). This aligns with the expected behavior of parameter-efficient fine-tuning: while adapters drastically reduce memory footprints and client download sizes, the forward and backward passes during training must still traverse the full frozen backbone, resulting in similar wall-clock computation times.

For full encoder-decoder adaptations, runtimes are naturally higher because both networks are optimized simultaneously. The adaptation time varies heavily depending on the target method's architectural complexity and the computational cost of its associated noise module. For instance, adapting to the heavier RoSteALS and WAM targets requires significantly more time compared to lighter CNN-based methods like HiDDeN, TrustMarkP, and InvisMark.

\begin{table*}[ht]
\centering
\caption{Training runtime analysis across different adaptation setups using TrustMarkQ as the base method. All adaptation variants are fine-tuned for 150 epochs on a single A100 (80GB) GPU, whereas the original TrustMarkQ base model is trained for 250 epochs.}
\label{tab:runtime}
\begin{tabular}{llcc}
\toprule
\textbf{Target Method} & \textbf{Adaptation Setup} & \textbf{Epochs} & \textbf{Total Time (Hours)} \\
\midrule
\textbf{TrustMarkQ} & Base Training (Original) & 250 & 59.6 \\
\midrule
\textbf{RoSteALS} & Encoder-Decoder & 150 & 110.5 \\
\textbf{WAM} & Encoder-Decoder & 150 & 71.9 \\
\textbf{TrustMarkP} & Encoder-Decoder & 150 & 44.5 \\
\textbf{InvisMark} & Encoder-Decoder & 150 & 40.9 \\
\textbf{HiDDeN} & Encoder-Decoder & 150 & 38.8 \\
\midrule
\textbf{RoSteALS} & Decoder-only (CWLoRA-16) & 150 & 30.6 \\
\textbf{WAM} & Decoder-only (CWLoRA-16) & 150 & 28.9 \\
\textbf{InvisMark} & Decoder-only (CWLoRA-16) & 150 & 26.8 \\
\textbf{TrustMarkP} & Decoder-only (CWLoRA-16) & 150 & 25.4 \\
\midrule
\textbf{RoSteALS} & Decoder-only (Full) & 150 & 30.8 \\
\textbf{WAM} & Decoder-only (Full) & 150 & 28.3 \\
\textbf{InvisMark} & Decoder-only (Full) & 150 & 26.1 \\
\textbf{TrustMarkP} & Decoder-only (Full) & 150 & 25.8 \\
\bottomrule
\end{tabular}
\end{table*}

% ====================================================================
% APPENDIX D: Configuration Details
% ====================================================================
\section{Configuration Details}
\label{sec:config_details}

\begin{table}[t]
\centering
\caption{Encoder and decoder parameter counts (in millions) and architecture type for each method. The CWLoRA-16 adapter parameters are reported with respect to TrustMarkQ as the \emph{base} method, matching the configuration used in our encoder–decoder adaptation experiments (Section~\ref{sec:exp_encdec}).}
\label{tab:wm_params}
\setlength{\tabcolsep}{8pt}
\renewcommand{\arraystretch}{1.2}

\resizebox{\columnwidth}{!}{%
\begin{tabular}{l l c l}
\toprule
\textbf{Method} &  & \textbf{Params (M)} & \textbf{Type} \\
\midrule
\multirow{2}{*}{TrustMark P/Q}
  & Encoder & 8.6  & CNN \\
  & Decoder & 23.7 & CNN \\
\addlinespace[2pt]
\multirow{2}{*}{HiDDeN}
  & Encoder & 0.2  & CNN \\
  & Decoder & 0.2  & CNN \\
\addlinespace[2pt]
\multirow{2}{*}{RoSteALS}
  & Encoder & 81.8 & VQGAN+CNN \\
  & Decoder & 23.7 & CNN \\
\addlinespace[2pt]
\multirow{2}{*}{InvisMark}
  & Encoder & 3.9  & CNN \\
  & Decoder & 87.6 & CNN (ConvNeXT) \\
\addlinespace[2pt]
\multirow{2}{*}{WAM}
  & Encoder & 1.1  & CNN+Attn+JND \\
  & Decoder & 93.3 & ViT \\
\addlinespace[2pt]
\midrule
\multirow{2}{*}{CWLoRA-16}
  & Encoder & 0.9  & CNN \\
  & Decoder & 2.0  & CNN \\
\bottomrule
\end{tabular}
}
\vspace{-0.4cm}
\end{table}

% \begin{table}[t]
% \centering
% \resizebox{1.0\textwidth}{!}{
% \begin{tabular}{ll cccccc}
% \toprule
% \multirow{2}{*}{\textbf{Component}} & \multirow{2}{*}{\textbf{Property}} & \multicolumn{6}{c}{\textbf{Method}} \\
% \cmidrule(lr){3-8}
% & & \textbf{TrustMark P/Q} & \textbf{HiDDeN} & \textbf{RoSteALS} & \textbf{InvisMark} & \textbf{WAM} & \textbf{CWLoRA-16} \\
% \midrule
% \multirow{2}{*}{Encoder} 
%  & Params (M) & 8.6 & 0.2 & 81.8 & 3.9 & 1.1 & 0.9 \\
%  & Type & CNN & CNN & VQGAN+CNN & CNN & CNN+Attn+JND & CNN \\
% \midrule
% \multirow{2}{*}{Decoder} 
%  & Params (M) & 23.7 & 0.2 & 23.7 & 87.6 & 93.3 & 2.0 \\
%  & Type & CNN & CNN & CNN & CNN (ConvNeXT) & ViT & CNN \\
% \bottomrule
% \end{tabular}
% }
% \caption{Encoder and decoder parameter counts (in millions) and architecture type for each method. The CWLoRA-16 adapter parameters are reported with respect to TrustMarkQ as the \emph{base} method, matching the configuration used in our encoder–decoder adaptation experiments (Section~\ref{sec:exp_encdec}).}
% \label{tab:wm_params}
% \end{table}

\subsection{Training Schedule}
\label{sec:training_schedule}

All adaptations run for 150 epochs with a batch size of 128 and learning rate $4\times10^{-5}$. Training proceeds in three phases: (1) a \emph{warm-up} on a fixed batch with no augmentation, switching to the full training set once bit accuracy reaches $0.90$; (2) continued training until accuracy reaches $0.95$; (3) activation of the target noise module $\mathcal{A}_\text{tgt}$ with strength linearly ramped to full over $\approx$16 epochs. For encoder--decoder adaptation, the secret-loss weights $\mu_2$ and $\lambda_1$ are ramped from 0 to 1 over 10k iterations, while the reconstruction weights $\alpha,\beta,\gamma,\delta$ are fixed at 1. Although all runs use 150 epochs to ensure convergence, experiments suggest much shorter training often causes only slight performance decreases.

\subsection{Noise Modules}
\label{sec:noise_modules}

In most of our experiments, when adapting to a target watermarking method, we use the corresponding noise module \(\mathcal{A}_\text{tgt}\), which is defined by the types and strengths of noises/augmentations reported for training that watermark (the only exception is HiDDeN, for which we use a lighter noise module due to its lower robustness). Each noise module applies a subset of its noises to every image according to a stochastic selection rule. The noise modules are used in the same way for both training and test.

We follow the noise application strategy of TrustMark (Q/P): among the available noise types, we treat VFlip, HFlip, Crop, and Rotation (when present in the module) as \textbf{fixed}, and all other noises as \textbf{optional}. For TrustMarkQ, TrustMarkP, and RoSteALS, the noise modules apply \emph{all} fixed noises and additionally sample two optional noises at random, which are then applied as well. For the remaining methods, which are generally less robust to combinations of noises because they were not trained with such mixtures, we use a milder configuration: for WAM, we apply one fixed and one optional noise to each image, while for InvisMark and HiDDeN we apply only a single noise (no mixture), sampled uniformly at random from the union of fixed and optional noises.

Below we list the noise types and strengths used for each method. Since TrustMarkQ is used as the base method in many of our experiments, using colors, we visualize the relative strength of the noises with respect to those used by TrustMarkQ. For new or stronger noises, we use the color \noisehigh{orange}, and for weaker noises, we use the color \noiselow{blue}.
\\

\noindent \textbf{TrustMarkQ}
\begin{lstlisting}[style=augstyle]
HFlip()
ResizedCrop(scale=(0.7,1.0), ratio=(0.75,1.33))
Compress(jpeg_quality=40)
Brightness(0.5,1.5)
Contrast(0.5,1.5)
GaussianBlur(kernel=7, sigma=(0.1,2.0))
Saturation(0.5,1.5)
GaussianNoise(std=0.08)
ColorJiggle(brightness=0.1, contrast=0.1, saturation=0.1, hue=0.05)
Posterize(bits=3)
RGBShift(limit=0.1)
BoxBlur(kernel=7) 
Hue(hue=0.05)
MedianBlur(kernel=3)
Grayscale()
MotionBlur(kernel=(3,9), angle=(-90,90), direction=(-1.0,1.0))
Sharpness(2.5)
\end{lstlisting}

\noindent \textbf{TrustMarkP}
\begin{lstlisting}[style=augstyle]
HFlip()
ResizedCrop(scale=(0.7,1.0), ratio=(0.75,1.33))
Compress(jpeg_quality=@\noiselow{70}@)
Brightness(@\noiselow{0.9,1.1}@)
Contrast(@\noiselow{0.9,1.1}@)
GaussianBlur(kernel=@\noiselow{3}@, sigma=@\noiselow{(0.1,1.0)}@)
Saturation(@\noiselow{0.9,1.1}@)
GaussianNoise(std=@\noiselow{0.02}@)
ColorJiggle(brightness=@\noiselow{0.05}@, contrast=@\noiselow{0.05}@, saturation=@\noiselow{0.05}@, hue=@\noiselow{0.01}@)
Posterize(bits=@\noiselow{5}@)
RGBShift(limit=@\noiselow{0.02}@)
BoxBlur(kernel=@\noiselow{3}@) 
Hue(hue=@\noiselow{0.01}@)
MedianBlur(kernel=3)
Grayscale()
MotionBlur(kernel=(3,@\noiselow{5}@), angle=@\noiselow{(-25,25)}@, direction=@\noiselow{(-0.25,0.25)}@)
Sharpness(@\noiselow{0.5}@)
\end{lstlisting}

\noindent \textbf{RoSteALS}
\begin{lstlisting}[style=augstyle]
Compress(jpeg_quality=@\noiselow{50}@)
Brightness(@\noiselow{0.7,1.3}@)
Contrast(0.5,@\noisehigh{2.0}@)
GaussianBlur(kernel=7, sigma=(@\noiselow{1.0},\noisehigh{3.0}@))
GaussianNoise(std=@\noiselow{0.02}@)
BoxBlur(kernel=@\noisehigh{7}@)
Hue(hue=@\noisehigh{0.1}@)
MedianBlur(kernel=3)
@\noisehigh{CustomSaturation()}@
@\noisehigh{RandomImagenetC()}@
\end{lstlisting}

\noindent \textbf{WAM}
\begin{lstlisting}[style=augstyle]
HFlip()
ResizedCrop(scale=(@\noisehigh{0.33}@,1.0), ratio=(0.75,1.33))
Compress(jpeg_quality=40)
Brightness(0.5,@\noisehigh{2.0}@)
Contrast(0.5,@\noisehigh{2.0}@)
GaussianBlur(kernel=@\noisehigh{(3,17)}@)
Saturation(0.5,@\noisehigh{2.0}@)
MedianBlur(kernel=@\noisehigh{(3,7)}@)
@\noisehigh{Perspective(distortion=0.5)}@
@\noisehigh{Rotate(-10,10)}@
\end{lstlisting}

\noindent \textbf{InvisMark}
\begin{lstlisting}[style=augstyle]
HFlip()
ResizedCrop(scale=(0.7,1.0), ratio=(0.75,1.33))
Compress(jpeg_quality=@\noiselow{50}@)
Brightness(@\noiselow{0.75,1.25}@)
Contrast(@\noiselow{0.75,1.25}@)
GaussianBlur(kernel=@\noiselow{5}@, sigma=(0.1,@\noiselow{1.5}@))
Saturation(@\noiselow{0.75,1.25}@)
GaussianNoise(std=@\noiselow{0.04}@)
ColorJiggle(brightness=0.1, contrast=0.1, saturation=0.1, hue=@\noiselow{0.02}@)
Posterize(bits=@\noiselow{4}@)
RGBShift(limit=@\noiselow{0.05}@)
BoxBlur(kernel=@\noiselow{5}@)
@\noisehigh{VFlip()}@
@\noisehigh{Rotate(0,10)}@
@\noisehigh{Erasing(scale=(0.02, 0.1), ratio=(0.5, 1.5))}@
@\noisehigh{Perspective(distortion=0.1)}@
\end{lstlisting}

\noindent \textbf{HiDDeN}
\begin{lstlisting}[style=augstyle]
ResizedCrop(scale=(0.7,1.0), ratio=(0.75,1.33))
Brightness(0.9,1.1)
\end{lstlisting}

\subsection{Adapter Details}
\label{sec:adapter_details}

Adapters are attached to CNN-based encoders/decoders. Unless noted, all existing convolutional weights are \emph{frozen} during training; adapter parameters and non-convolutional layers (e.g., batch norm, FC) remain trainable. For a convolutional layer $\ell$ with kernel
$W^{(\ell)} \!\in\! \mathbb{R}^{C^{(\ell)}_{\text{out}}\times C^{(\ell)}_{\text{in}}\times K^{(\ell)}\times K^{(\ell)}}$,
each adapter below modifies the forward or injects a low-rank update as follows.

\paragraph{\textbf{CWLoRA-$r$.}}
Adds a rank-$r$ weight-space update via two sparse tensors
$A^{(\ell)} \!\in\! \mathbb{R}^{r\times C^{(\ell)}_{\text{in}}\times K^{(\ell)}\times K^{(\ell)}}$ and
$B^{(\ell)} \!\in\! \mathbb{R}^{r\times C^{(\ell)}_{\text{out}}\times K^{(\ell)}\times K^{(\ell)}}$:
\begin{equation*}
\begin{aligned}
\Delta W^{(\ell)}_{o,i,u,v} \;=\;  & \alpha_{\text{cw}}\sum_{t=1}^{r} B^{(\ell)}_{t,o,u,v}\,A^{(\ell)}_{t,i,u,v}, \\
\tilde{W}^{(\ell)} \;=\; & W^{(\ell)} + \Delta W^{(\ell)}.
\end{aligned}
\end{equation*}

Only $A^{(\ell)}$ and $B^{(\ell)}$ are trainable; no extra inference ops are introduced (the update is merged into weights).

\paragraph{\textbf{PEFT-LoRA-$r$.}}
Following PEFT's Conv2d LoRA, each \(K{\times}K\) convolution is wrapped with two trainable adapters composed as
\[
\Delta y^{(\ell)} \;=\; \tfrac{\alpha_{\text{peft}}}{r}\;\underbrace{\mathrm{Conv}_{K\times K}^{(\ell)}\!\big( \underbrace{\mathrm{Conv}_{1\times 1}^{(\ell)}(x)}_{\text{reduce } C_{\text{in}}\!\to r} \big)}_{\text{expand } r\!\to C_{\text{out}}}\!,
\]
and the forward becomes
\[
\tilde{y}^{(\ell)} \;=\; \mathrm{Conv}_{K\times K}(x; W^{(\ell)}, b^{(\ell)}) \;+\; \Delta y^{(\ell)}.
\]
Here the \(1{\times}1\) \(\mathrm{Conv}_{1\times 1}^{(\ell)}\) reduces channels from \(C_{\text{in}}\) to rank \(r\), and the \(K{\times}K\) \(\mathrm{Conv}_{K\times K}^{(\ell)}\) restores to \(C_{\text{out}}\) while inheriting the base layer's stride, padding, dilation, and groups.
The base conv weights/bias are frozen; only the LoRA adapters are trainable. In our experiments we set \(\alpha_{\text{peft}} = 4r\).

\paragraph{\textbf{IA$^3$.}}
Freezes conv weights and learns a per-channel scaling vector applied either to inputs or outputs (default: outputs). For $y\!=\!\mathrm{Conv}(x;W,b)$ and a trainable $s_{\text{out}}\!\in\!\mathbb{R}^{C_{\text{out}}}$,
\begin{equation*}
\begin{aligned}
\tilde{y} \;=\; & y \odot s_{\text{out}}\quad\text{(output gating)}\qquad 
\text{or} \\
\tilde{y} \;=\; &  \mathrm{Conv}(x\odot s_{\text{in}};W,b)\ \text{(input gating)}.
\end{aligned}
\end{equation*}
This introduces only $C_{\text{out}}$ (or $C_{\text{in}}$) parameters per conv; biases remain trainable.

\paragraph{\textbf{$1\times 1$-Bottleneck (rank $r$).}}
Adds a residual adapter on the conv output:
\begin{equation*}
\begin{aligned}
y \;=\; &  \mathrm{Conv}(x;W,b), \quad
\Delta y \;=\; \mathrm{Up}\!\big(\sigma(\mathrm{Down}(y))\big), \\
\tilde{y} \;=\; & y + \alpha_{\text{cba}}\,\Delta y,
\end{aligned}
\end{equation*}
where $\mathrm{Down}\!:\mathbb{R}^{C_{\text{out}}}\!\to\!\mathbb{R}^{r}$ and
$\mathrm{Up}\!:\mathbb{R}^{r}\!\to\!\mathbb{R}^{C_{\text{out}}}$ are $1{\times}1$ convolutions and $\sigma$ is ReLU. The base conv is frozen; the two $1{\times}1$ layers and activation are trainable.

\paragraph{\textbf{SE-Gate.}}
Implements a squeeze–excitation gate on the conv output while freezing the base conv:
\begin{equation*}
\begin{aligned}
y \;=\; & \mathrm{Conv}(x;W,b), \\
s \;=\; & \sigma\!\big(W_2\,\mathrm{ReLU}(W_1\,\mathrm{GAP}(y))\big), \\
\tilde{y} \;=\; & y \odot s,
\end{aligned}
\end{equation*}
with $\mathrm{GAP}$ global average pooling, $W_1\!:\mathbb{R}^{C_{\text{out}}}\!\to\!\mathbb{R}^{\max(4,\lfloor C_{\text{out}}/32\rfloor)}$, and $W_2$ projecting back to $C_{\text{out}}$. Only $W_1,W_2$ are trainable.

\paragraph{\textbf{CWLoRA-$r$ for ConvNeXt (InvisMark decoder).}}
ConvNeXt couples depthwise convolutions with MLP blocks. Our ConvNeXt-specific variant leaves depthwise \texttt{Conv2d} layers \emph{trainable} (no LoRA), excludes classifier/head layers (kept fully trainable), and installs budgeted LoRA on \texttt{Linear} layers inside blocks:
for $z\!=\!\mathrm{Linear}(u;W)$, we use
\[
\tilde{W} \;=\; W + \tfrac{\alpha_\text{cnext}}{r}\,A B^\top,\qquad
A\!\in\!\mathbb{R}^{d_{\text{in}}\times r},\ B\!\in\!\mathbb{R}^{r\times d_{\text{out}}},
\]
with a learned gate and per-layer ranks chosen to meet a global trainable-parameter target. Defaults: target ratio $20\%$ of total params, $\alpha\!=\!16$, per-layer ranks $r\!\in\![8,128]$ (expand/projection layers prioritized), classifier/head excluded. Under these settings, the adapter has $\approx\!19$M trainable parameters on a 102M-parameter base decoder.

\medskip\noindent
\textit{Notes.} In Table~\ref{tab:base-decoder-comparison}, the same CWLoRA implementation is used across base decoders. InvisMark's ConvNeXt-based decoder requires the ConvNeXt-specific variant above due to its depthwise+MLP design.

% ====================================================================
% APPENDIX E: Multi-Decoder Inference Details
% ====================================================================
\section{Multi-Decoder Inference Details}

\subsection{Inference and Selection Pseudo-Codes}
\label{sec:multi_dec_inf_pseudo}

\paragraph{Setting and notation.}
We assume an image \(I\) has already been watermarked by \emph{one} method with a codeword of length \(s\) that encodes data bits \(d \in \{0,1\}^{L}\) using a BCH codec \(\mathrm{ECC}\) (so \(L=\mathrm{ECC.data\_length}(s)\)).
Let \(\mathcal{M}=\{(D_k, s_k)\}_{k=1}^K\) denote the set of available watermark methods used at inference time, where \(D_k\) is the decoder and \(s_k\) its secret length (original or adapterized, both handled identically here).
Given \(I\), the goal is to (i) select the correct decoder and (ii) recover the embedded data bits.

\paragraph{Selection policies.}
We study two ways to select which decoder to run:
(i) \emph{ECC-based} (plug-and-play): sequentially try decoders in a random order and return the first whose ECC verification succeeds; and
(ii) \emph{classifier-based:} a small classifier predicts the watermark method, after which only the predicted decoder is used and its output is ECC-decoded.
An oracle that knows the true method provides an upper bound.

The pseudo-codes for the selection and evaluation procedures are provided below.

\begin{algorithm}[t]
\caption{Decode with selection (per image)}
\label{alg:decode-with-selection}
\begin{algorithmic}[1]
\Require image \(I\); methods \(\mathcal{M}=\{(D_k,s_k)\}_{k=1}^K\); selector \(\mathsf{Select}\); ECC codec
\Ensure selected method \(\hat{k}\), recovered data bits \(\hat{d}\), verification flag \(\mathsf{ok}\)
\State \((\hat{k}, \hat{d}, \mathsf{ok}) \gets \mathsf{Select}(I,\mathcal{M},\mathrm{ECC})\)
\State \Return \((\hat{k}, \hat{d}, \mathsf{ok})\)
\end{algorithmic}
\end{algorithm}

\begin{algorithm}[t]
\caption{Per-image evaluation (requires ground truth)}
\label{alg:evaluate-one}
\begin{algorithmic}[1]
\Require image \(I\) carrying ground-truth data bits \(d\); true method index \(k^\star\); \(\mathcal{M}\); \(\mathrm{ECC}\); selector \(\mathsf{Select}\)
\Ensure FA (false assignment), FD (false decoding), ACC (correct full decode)
\State $(\hat{k}, \hat{d}, \mathsf{ok}) \gets \textsc{DecodeWithSelection}(I, M,$
\Statex \hspace{\algorithmicindent} $\mathsf{Select}, \mathrm{ECC})$
\If{\(\hat{k} \neq k^\star\)} \State \Return FA\(=1\), FD\(=0\), ACC\(=0\)
\ElsIf{\(\mathsf{ok}=\text{false}\ \lor\ \hat{d} \neq d\)} \State \Return FA\(=0\), FD\(=1\), ACC\(=0\)
\Else \State \Return FA\(=0\), FD\(=0\), ACC\(=1\)
\EndIf
\end{algorithmic}
\end{algorithm}

\begin{algorithm}[t]
\caption{ECC-based selection}
\label{alg:select-ecc}
\begin{algorithmic}[1]
\Function{SelectECC}{$I,\mathcal{M},\mathrm{ECC}$}
  \State \(\pi \gets\) a random permutation of \(\{1,\dots,K\}\) \Comment{shuffle decoders}
  \For{each \(k \in \pi\)}
     \State \(\hat{c} \gets D_k(I)\) \Comment{binary codeword \(\hat{c}\in\{0,1\}^{s_k}\)}
     \State \((\hat{d}, \mathsf{ok}) \gets \mathrm{ECC.decode}(\hat{c}, s_k)\)
     \If{\(\mathsf{ok}\)} \State \Return \((k,\hat{d},\text{true})\)
     \EndIf
  \EndFor
  \State \Return \((\text{None},\text{None},\text{false})\)
\EndFunction
\end{algorithmic}
\end{algorithm}

\begin{algorithm}[t]
\caption{Classifier-based selection}
\label{alg:select-clf}
\begin{algorithmic}[1]
\Function{SelectClf}{$I,\mathcal{M},\mathrm{ECC},\mathcal{C}$}
  \State \(\hat{k} \gets \arg\max \mathcal{C}(I)\)
  \State \(\hat{c} \gets D_{\hat{k}}(I)\) \Comment{binary codeword}
  \State \((\hat{d}, \mathsf{ok}) \gets \mathrm{ECC.decode}(\hat{c}, s_{\hat{k}})\)
  \State \Return \((\hat{k}, \hat{d}, \mathsf{ok})\)
\EndFunction
\end{algorithmic}
\end{algorithm}

\subsection{Error Rates}
\label{sec:multi_dec_errors}

\begin{table*}[t]
\centering
\caption{\textbf{False-assignment and false-decoding rates for multi-decoder inference.} Same runs as Table~\ref{tab:multi-dec}. FA = false assignment (wrong decoder), FD = false decoding (right decoder, but ECC or data bits fail), Acc = 100 - FA - FD.}
\label{tab:multi-dec-errors}
\setlength{\tabcolsep}{4pt}
\resizebox{\textwidth}{!}{%
\begin{tabular}{lcccccccc}
\toprule
\multirow{2}{*}{Method} &
\multicolumn{3}{c}{Clean} &
\multicolumn{3}{c}{Noised} &
\multirow{2}{*}{Model Size (GB)} &
\multirow{2}{*}{Runtime (ms)}
\\
\cmidrule(lr){2-4} \cmidrule(lr){5-7}
& FA (\%) & FD (\%) & Acc (\%) & FA (\%) & FD (\%) & Acc (\%) & & \\
\midrule
Original+Oracle      & 0.00 & 3.30 & 96.70 & 0.00 & 13.83 & 86.17 & 1.19 & 1.35  \\
Adapter+Oracle       & 0.00 & 2.77 & 97.23 & 0.00 & 13.81 & 86.19 & 0.14 & 0.62 \\
\midrule
Original+Classifier  & 0.48 & 3.22 & 96.30 & 2.39 & 11.95 & 85.66 & 1.29 & 1.79 \\
Adapter+Classifier   & 0.48 & 2.65 & \textbf{96.87} & 2.28 & 11.99 & \textbf{85.73} & 0.24 & \textbf{1.07} \\
Original+ECC         & 3.86 & 0.01 & 96.13 & 14.25 & 0.07 & 85.68 & 1.19  & 4.75\\
Adapter+ECC          & 4.44 & 0.01 & 95.55 & 15.26 & 0.08 & 84.66 & \textbf{0.14} & 2.18 \\
\bottomrule
\end{tabular}%
}
\end{table*}

Table~\ref{tab:multi-dec-errors} reports false assignment and false decoding for the same runs as Table~\ref{tab:multi-dec}.

\subsection{Per-Watermark Performance}
\label{sec:multi_dec_inf_perwm}

\begin{figure*}[t]
    \centering
    \includegraphics[width=1.0\linewidth]{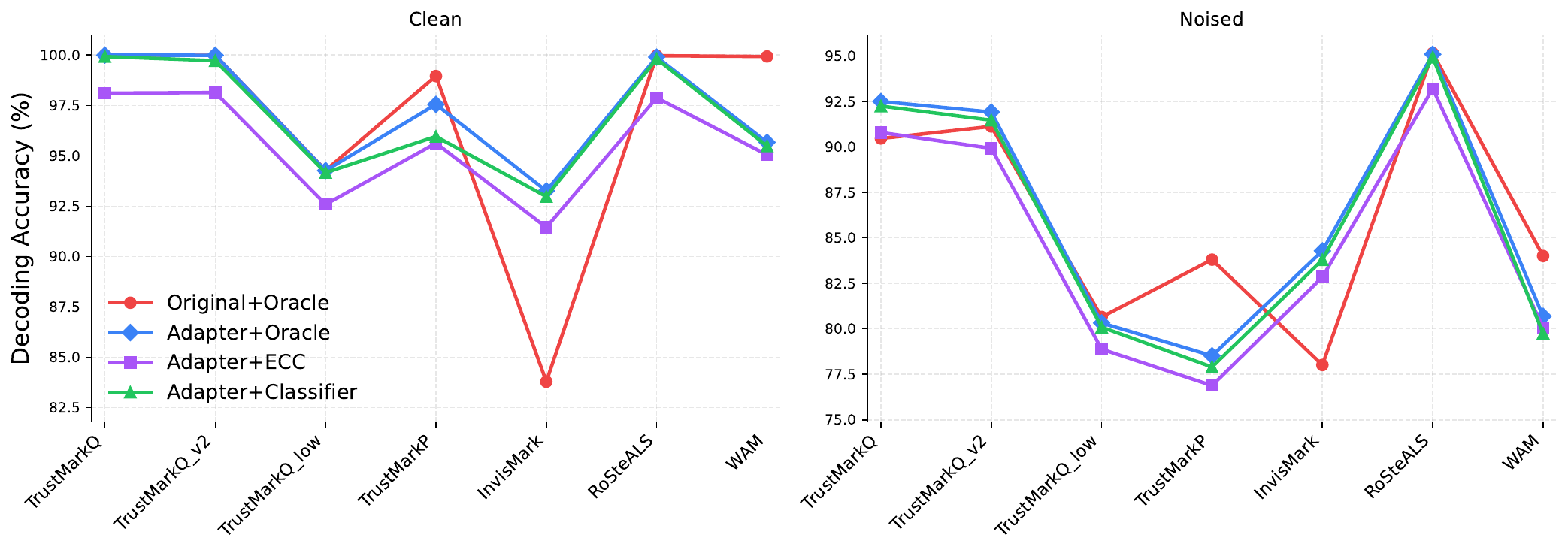}
    \caption{\textbf{Per-watermark comparison of approaches for multi-decoder inference.} Comparison of multi-decoder framework implemented with different settings on images watermarked by each watermarking method.}
    \label{fig:multi-dec-per-method}
\end{figure*}

Figure~\ref{fig:multi-dec-per-method} shows the decoding accuracy of different multi-decoder framework algorithms on images watermarked by each watermarking method. This uses the same setting as Table~\ref{tab:multi-dec}, but instead of the average overall decoding accuracy across all watermarking methods, per-watermark performance is shown.

\subsection{Classifier Training}
\label{sec:selection_classifier_train}

The classifier in Section~\ref{sec:multi-dec} that is used for decoder selection has a ResNet50 architecture, and is trained to distinguish between images watermarked by different methods. The training uses the MIRFLICKR dataset with 80k images, and is done for 10 epochs with Adam optimizer and learning-rate of $10^{-4}$. After achieving overall classification accuracy of 98\% on clean images, the specialized noise modules for each watermarking method are activated. The final model achieves 98.23\% test accuracy on noised images.

\section{Additional Watermark Residual Visualizations}
\label{sec:adaptation_visualizations}

\begin{figure*}[t]
\centering
\setlength{\tabcolsep}{2pt}

\resizebox{\textwidth}{!}{%
% 7 Columns total. 'm' columns ensure perfect vertical centering for images.
% Col 1: Inputs
% Col 2: Rotated Labels (with custom hskip to separate them visually)
% Col 3-7: The 5 Methods
\newcommand{\thead}[1]{\makebox[\linewidth][c]{\scriptsize\textbf{#1}}}
\begin{tabular}{ >{\centering\arraybackslash}m{0.14\textwidth} @{\hskip 16pt} c @{\hskip 10pt} *{5}{>{\centering\arraybackslash}m{0.14\textwidth}} }

% ==================== HEADER ROW ====================
& 
& \thead{TrustMarkP} 
& \thead{RoSteALS} 
& \thead{InvisMark} 
& \thead{HiDDeN} 
& \thead{WAM} \\[3pt]

% ==================== IMAGE 190314 ====================

% --- Row 1: Target Images ---
\includegraphics[width=\linewidth]{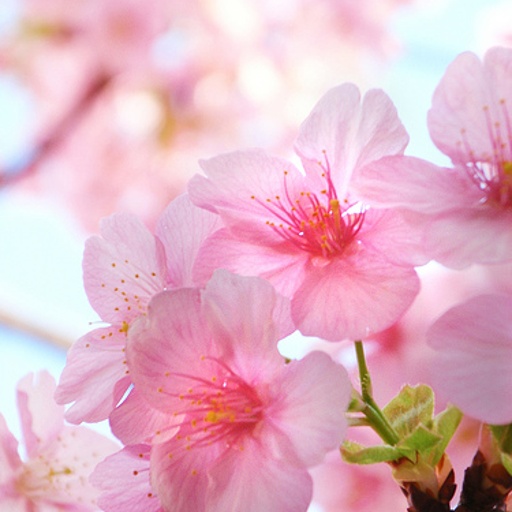} &
\raisebox{-0.4\height}{\rotatebox{90}{\textbf{Target}}} &
\includegraphics[width=\linewidth]{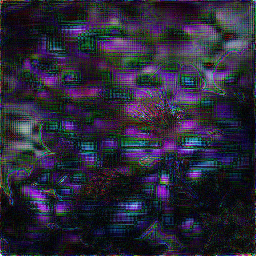} &
\includegraphics[width=\linewidth]{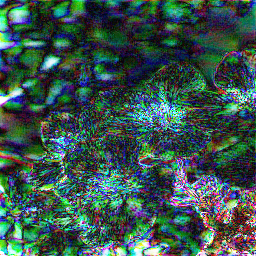} &
\includegraphics[width=\linewidth]{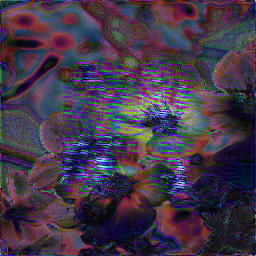} &
\includegraphics[width=\linewidth]{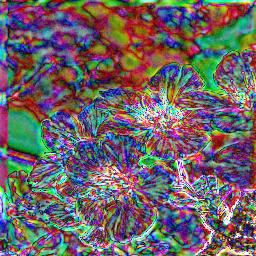} &
\includegraphics[width=\linewidth]{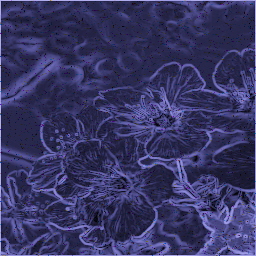} \\[-3pt]

% --- Row 2: Text below Col 1 only ---
{\scriptsize Original} & & & & & & \\[2pt]

% --- Row 3: Adapted Images ---
\includegraphics[width=\linewidth]{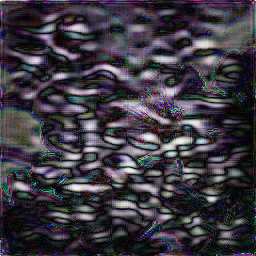} &
\raisebox{-0.45\height}{\rotatebox{90}{\textbf{Adapted}}} &
\includegraphics[width=\linewidth]{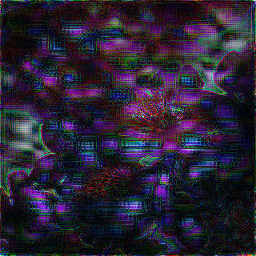} &
\includegraphics[width=\linewidth]{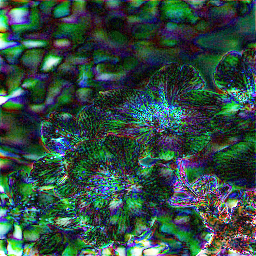} &
\includegraphics[width=\linewidth]{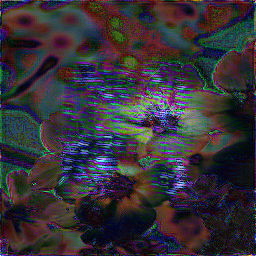} &
\includegraphics[width=\linewidth]{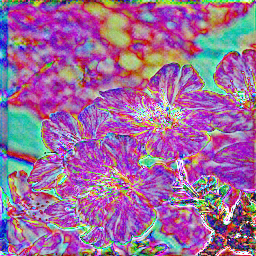} &
\includegraphics[width=\linewidth]{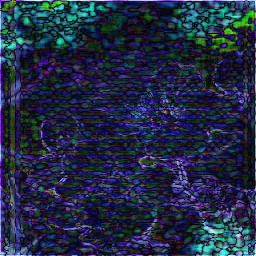} \\[-3pt]

% --- Row 4: Text below Col 1 only ---
\begin{tabular}{@{}c@{}}\scriptsize Base residual \\[-2pt] \scriptsize (TrustMarkQ)\end{tabular} & & & & & & \\[12pt]

% ==================== IMAGE 190428 ====================

% --- Row 5: Target Images ---
\includegraphics[width=\linewidth]{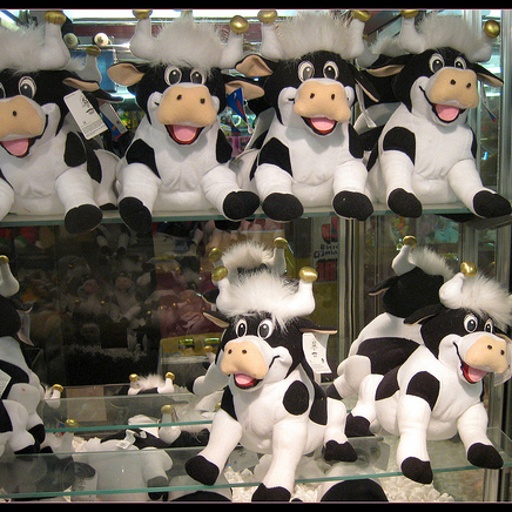} &
\raisebox{-0.4\height}{\rotatebox{90}{\textbf{Target}}} &
\includegraphics[width=\linewidth]{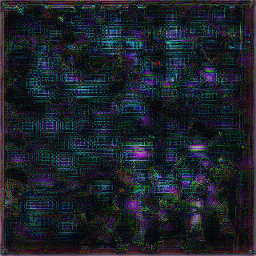} &
\includegraphics[width=\linewidth]{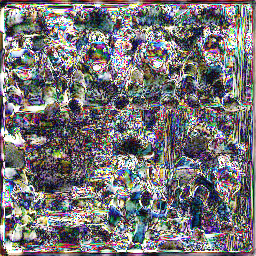} &
\includegraphics[width=\linewidth]{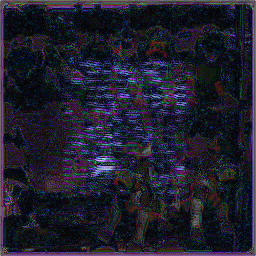} &
\includegraphics[width=\linewidth]{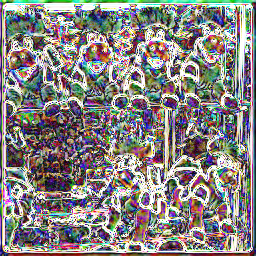} &
\includegraphics[width=\linewidth]{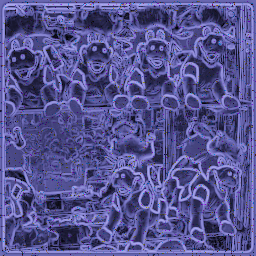} \\[-3pt]

% --- Row 6: Text below Col 1 only ---
{\scriptsize Original} & & & & & & \\[2pt]

% --- Row 7: Adapted Images ---
\includegraphics[width=\linewidth]{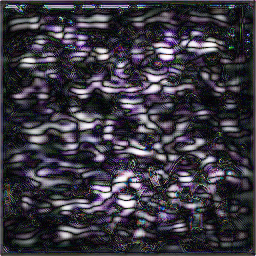} &
\raisebox{-0.4\height}{\rotatebox{90}{\textbf{Adapted}}} &
\includegraphics[width=\linewidth]{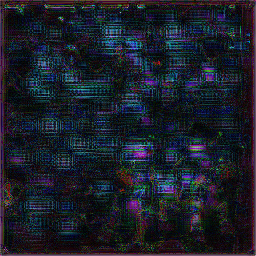} &
\includegraphics[width=\linewidth]{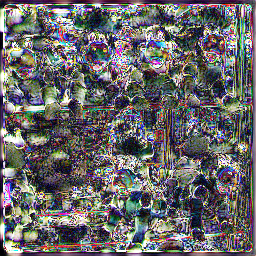} &
\includegraphics[width=\linewidth]{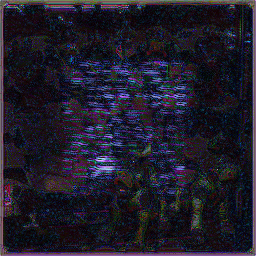} &
\includegraphics[width=\linewidth]{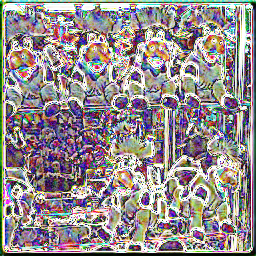} &
\includegraphics[width=\linewidth]{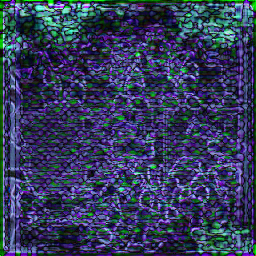} \\[-3pt]

% --- Row 8: Text below Col 1 only ---
\begin{tabular}{@{}c@{}}\scriptsize Base residual \\[-2pt] \scriptsize (TrustMarkQ)\end{tabular} & & & & & & \\[12pt]

% ==================== IMAGE 190317 ====================

% --- Row 9: Target Images ---
\includegraphics[width=\linewidth]{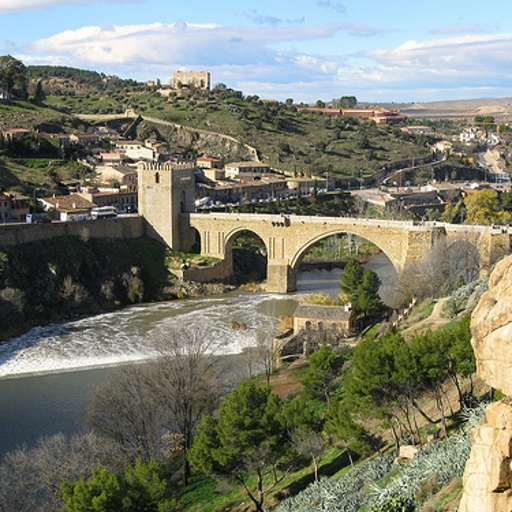} &
\raisebox{-0.4\height}{\rotatebox{90}{\textbf{Target}}} &
\includegraphics[width=\linewidth]{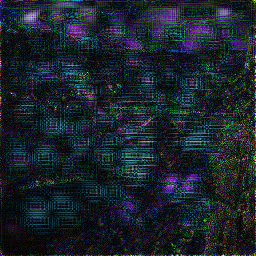} &
\includegraphics[width=\linewidth]{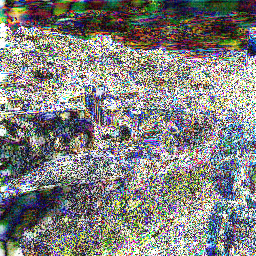} &
\includegraphics[width=\linewidth]{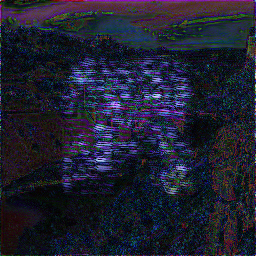} &
\includegraphics[width=\linewidth]{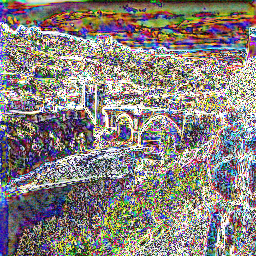} &
\includegraphics[width=\linewidth]{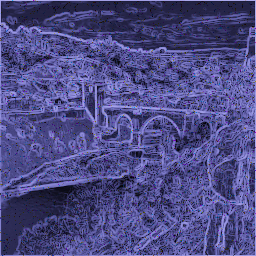} \\[-3pt]

% --- Row 10: Text below Col 1 only ---
{\scriptsize Original} & & & & & & \\[2pt]

% --- Row 11: Adapted Images ---
\includegraphics[width=\linewidth]{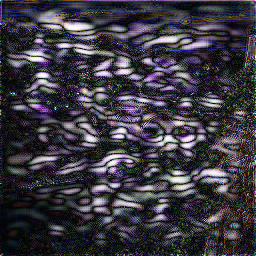} &
\raisebox{-0.4\height}{\rotatebox{90}{\textbf{Adapted}}} &
\includegraphics[width=\linewidth]{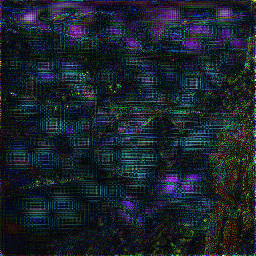} &
\includegraphics[width=\linewidth]{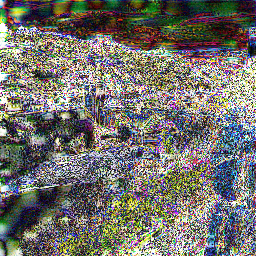} &
\includegraphics[width=\linewidth]{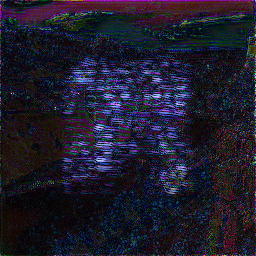} &
\includegraphics[width=\linewidth]{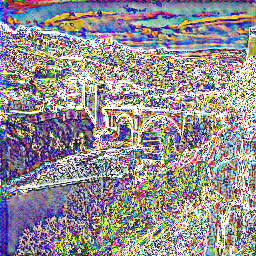} &
\includegraphics[width=\linewidth]{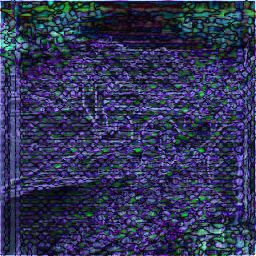} \\[-3pt]

% --- Row 12: Text below Col 1 only ---
\begin{tabular}{@{}c@{}}\scriptsize Base residual \\[-2pt] \scriptsize (TrustMarkQ)\end{tabular} & & & & & & \\

\end{tabular}
}% end resizebox

\vspace{0.2cm}
\caption{\textbf{Additional residual alignment examples for encoder adaptation.} Images are from validation set of MIRFLICKR.}
\label{fig:residual_grid_extra}
\end{figure*}

Figure~\ref{fig:residual_grid} visualizes watermark residuals produced by \(\Eft\) and \(\Etarget\) on a sample image.
The \(\Eft\) residuals become visually aligned with \(\Etarget\)'s while shifting significantly from \(\Ebase\), achieved with only lightweight updates to \(\Ebase\).
Figure~\ref{fig:residual_grid_extra} shows the same alignment on more sample images.

\end{document}